\documentclass[letterpaper]{article} % DO NOT CHANGE THIS
\usepackage[draft]{aaai2026}  % DO NOT CHANGE THIS
\usepackage{times}  % DO NOT CHANGE THIS
\usepackage{helvet}  % DO NOT CHANGE THIS
\usepackage{courier}  % DO NOT CHANGE THIS
\usepackage[hyphens]{url}  % DO NOT CHANGE THIS
\usepackage{graphicx} % DO NOT CHANGE THIS
\usepackage{natbib}  % DO NOT CHANGE THIS AND DO NOT ADD ANY OPTIONS TO IT
\usepackage{caption} % DO NOT CHANGE THIS AND DO NOT ADD ANY OPTIONS TO IT
\usepackage{algorithm}
\usepackage{algorithmic}
\usepackage{booktabs}       % professional-quality tables
\usepackage{amsmath}
\usepackage{amssymb}
\usepackage{amsfonts}
\usepackage{mathtools}
\usepackage{amsthm}
\usepackage{nicefrac}       % compact symbols for 1/2, etc.
\usepackage{microtype}      % microtypography
\usepackage{xcolor}         % colors
\usepackage{graphicx}       % figures
\usepackage{subcaption}
\usepackage{enumitem}
\usepackage{multirow}

\usepackage{newfloat}
\usepackage{listings}
\DeclareCaptionStyle{ruled}{labelfont=normalfont,labelsep=colon,strut=off} % DO NOT CHANGE THIS
\floatstyle{ruled}
\newfloat{listing}{tb}{lst}{}
\floatname{listing}{Listing}
\title{Planning in Branch-and-Bound: Model-Based Reinforcement Learning  \\ for Exact Combinatorial Optimization}
\author {
    Paul Strang\textsuperscript{\rm 1, \rm 3},
    Zacharie Alès\textsuperscript{\rm 2},
    Côme Bissuel\textsuperscript{\rm 1}, 
    Olivier Juan\textsuperscript{\rm 1},\\
    Safia Kedad-Sidhoum\textsuperscript{\rm 3},
    Emmanuel Rachelson\textsuperscript{\rm 4}
}
\affiliations {
    \textsuperscript{\rm 1}EDF R\&D, France\\
    \texttt{\{paul.strang, come.bissuel, olivier.juan\}@edf.fr} \\
    \textsuperscript{\rm 2}ENSTA Paris, Institut Polytechnique de Paris, France \\
    \texttt{zacharie.ales@ensta.fr} \\
    \textsuperscript{\rm 3}CNAM Paris, CEDRIC, France\\
    \texttt{safia.kedad\_sidhoum@cnam.fr} \\
    \textsuperscript{\rm 4}ISAE-SUPAERO, France\\
    \texttt{emmanuel.rachelson@isae-supaero.fr} \\
}

\newcommand{\greent}[1]{\textcolor{black}{#1}}
\newcommand{\greeng}[1]{\textcolor{black}{#1}}

\begin{document}
\maketitle
\begin{abstract}
    Mixed-Integer Linear Programming (MILP) lies at the core of many real-world combinatorial optimization (CO) problems, traditionally solved by branch-and-bound (B\&B). 
    A key driver influencing B\&B solvers efficiency is the variable selection heuristic that guides branching decisions. 
    Looking to move beyond static, hand-crafted heuristics, recent work has explored adapting traditional reinforcement learning (RL) algorithms to the B\&B setting, aiming to learn branching strategies tailored to specific MILP distributions. 
    In parallel, RL agents have achieved remarkable success in board games, a very specific type of combinatorial problems, by leveraging environment simulators to plan via Monte Carlo Tree Search (MCTS).
    Building on these developments, we introduce Plan-and-Branch-and-Bound (PlanB\&B), a model-based reinforcement learning (MBRL) agent that leverages a learned internal model of the B\&B dynamics to discover improved branching strategies. 
    Computational experiments empirically validate our approach, with our MBRL branching agent outperforming previous state-of-the-art RL methods across four standard MILP benchmarks. 
\end{abstract}

\section{Introduction}
    Mixed-Integer Linear Programming (MILP) plays a central role in combinatorial optimization (CO), a discipline concerned with finding optimal solutions over typically large but finite sets. 
    Specifically, MILPs offer a general modeling framework for NP-hard problems, and have become indispensable in tackling complex decision-making tasks across fields as diverse as operations research \citep{hillier2015introduction}, quantitative finance \citep{mansini2015linear}, and computational biology \citep{gusfield2019integer}.
    Modern MILP solvers are built upon the branch-and-bound (B\&B) paradigm \citep{land1960automatic}, which systematically explores the solution space by recursively partitioning the original problem into smaller subproblems, while maintaining provable optimality guarantees. 
    Since the 1980s, considerable research and engineering effort have gone into refining these solvers, resulting in highly optimized systems driven by expert-designed heuristics tuned over large benchmarks \citep{bixby_brief_2012, miplib_2021}.
    Nevertheless, in operational settings where structurally similar problems are solved repeatedly, adapting solver heuristics to the distribution of encountered MILPs can lead to substantial gains in efficiency, beyond what static, hand-crafted heuristics can offer.
    % efficient, data-driven 
    Recent research has thus turned to machine learning (ML) to design efficient, data-driven B\&B heuristics tailored to specific instance distributions \citep{scavuzzo_machine_2024}. 
    The variable selection heuristic, or branching heuristic, plays a particularly critical role in B\&B overall computational efficiency \citep{achterberg2013mixed}, as it governs the selection of variables along which the search space is recursively split.
    A key milestone was achieved by \citet{gasse_exact_2019}, who first managed to outperform human-expert branching heuristics by learning to replicate the behaviour of a greedy branching expert at lower computational cost. 
    While subsequent works succeeded in learning efficient branching strategies by reinforcement \citep{etheve_reinforcement_2020, scavuzzo_learning_2022}, none have yet matched the performance achieved by imitation learning (IL) approaches.
    This trend extends beyond MILPs to combinatorial optimization problems at large, as reinforcement learning (RL) baselines consistently underperform both handcrafted heuristics and IL methods trained to replicate expert strategies across various CO benchmarks \citep{berto_rl4co_2023}.
    Yet, if the performance of IL heuristics is capped by that of the experts they learn from, the performance of RL agents is, in theory, only bounded by the maximum score achievable. 
    One of the main challenges lies in the high dimensionality and combinatorial complexity of CO problems, which exacerbates exploration and credit assignment issues in sequential decision making.
    While supervised learning can partially mitigate these issues by scaling up neural architectures and exploiting abundant labeled data, such approaches are impractical in RL due to sample inefficiency and unstable learning dynamics \citep{liu_promoting_2023}.
    
\begin{figure}
        \centering
        \includegraphics[trim={0pt 0pt 0pt 0pt}, clip, width=0.45\textwidth]{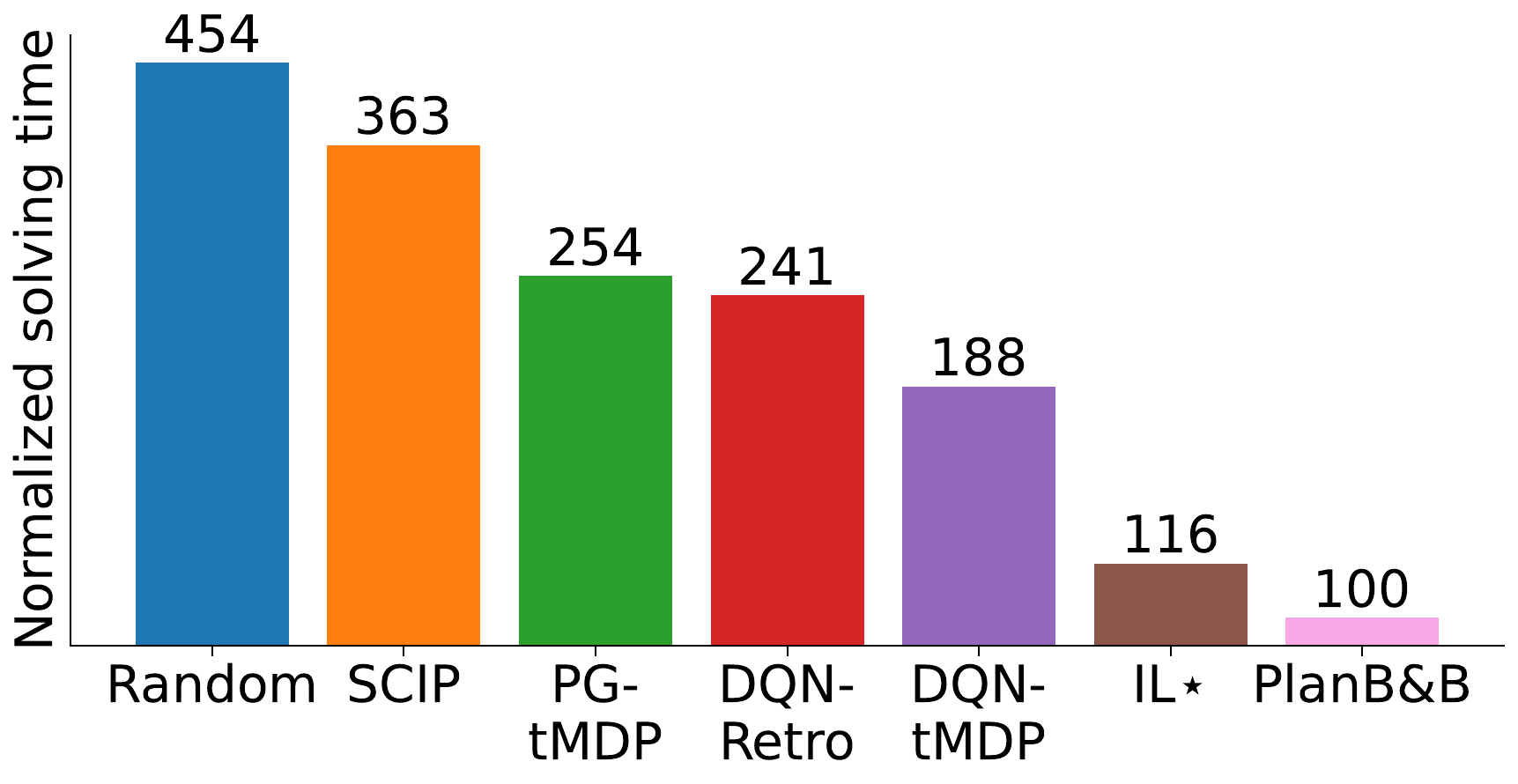}
      \caption{Aggregate normalized solving time performance obtained on test instances by SCIP \citep{bestuzheva_scip_2021}, IL, RL and random baselines across the Ecole benchmark \citep{prouvost_ecole_2020}, in log scale. These baselines are formally introduced in Section \ref{sec:exp_setup}.}
      \label{fig:teasing}
    \end{figure}

    Despite these challenges, RL has achieved notable success in a specific subset of combinatorial problems, board games, reaching superhuman performance by leveraging environment simulators to perform model-based planning. 
    Specifically, the integration of learned policy and value functions with look-ahead search via Monte Carlo Tree Search (MCTS) was found to steer action selection towards high-value states, effectively mitigating exploration and credit assignment issues in sparse-reward environment settings. 
    Inspired by the work of \citet{schrittwieser_mastering_2020}, we seek to extend the applicability of MCTS-based RL algorithms from board games to exact combinatorial optimization. To that end, we introduce Plan-and-Branch-and-Bound (PlanB\&B), a model-based reinforcement learning (MBRL) agent that leverages an internal model of the B\&B dynamics to learn improved variable selection strategies. To the best of our knowledge, this is the first MBRL agent specifically designed to solve CO problems. As shown in Figure \ref{fig:teasing}, our agent achieves state-of-the-art performance on test sets across four standards MILP benchmarks \citep{prouvost_ecole_2020}, surpassing prior IL and RL baselines. These results suggest that the branching dynamics in B\&B can be approximated with sufficient fidelity to enable policy improvement through planning over a learned model, opening the door to broader applications of MBRL to mixed-integer linear programming.

\section{Problem statement} 
    \paragraph{Mixed-integer linear programming.}
    \label{sec:b&b}
    \greeng{We consider mixed-integer linear programs (MILPs), defined as:}
        \[ P : 
            \left\{
                \begin{array}{ll}
                    \min c^\top x \\
                    l \leq x \leq u \\
                     Ax \leq b \; ; \; x \in \mathbb{Z}^{|\mathcal{I}|} \times \mathbb{R}^{n-|\mathcal{I}|} \\

                \end{array}
            \right.
        \]
    
    \greeng{with~$n$ the number of variables,~$m$ the number of linear constraints,~$l, u \in \mathbb{R}^n$ the lower and upper bound vectors,~$A\in \mathbb{R}^{m\times n}$ the constraint matrix,~$b\in \mathbb{R}^m$ the right-hand side vector,~$c\in \mathbb{R}^n$ the objective function, and~$\mathcal{I}$ the indices of integer variables. In this work, we are interested in repeated MILPs of fixed dimension~$\{P_{i} = (A_i, b_i, c_i, l_i, u_i)\}_{i\in \mathcal{D}}$ sampled according to an unknown distribution~$p_0$.}
    
\begin{figure*}[t!]
        \centering
        \includegraphics[width=1\textwidth]{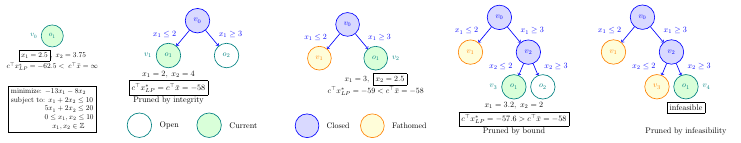}
        \caption{\greeng{Solving a MILP by B\&B using variable selection policy $\pi$ and node selection policy $\rho$. Each node $v_i$ represents a MILP derived from the original problem, each edge represents the bound adjustment applied to derive child nodes from their parent. At each step, nodes $o_i \in \mathcal{O}$ are re-indexed according to $\rho$.}}
        \label{fig:BB}
    \end{figure*}

    In order to solve MILPs efficiently, the B\&B algorithm iteratively builds a binary tree $(\mathcal{V}, \mathcal{E}) $ where each node corresponds to a MILP, starting from the root node $v_0 \in \mathcal{V}$ representing the original problem $P_0$. 
    The incumbent solution $\bar{x} \in \mathbb{Z}^{|\mathcal{I}|} \times \mathbb{R}^{n-|\mathcal{I}|}$ denotes the best feasible solution found at current iteration, its associated value $GU\!B = c^\top \bar{x}$ is called the \textit{global upper bound} on the optimal value. 
    The overall state of the optimization process is thus captured by the triplet $s = (\mathcal{V}, \mathcal{E}, \bar{x})$, we note $\mathcal{S}$ the set of all such triplets.
    Throughout the optimization process, B\&B nodes are explored sequentially. We note $\mathcal{C}$ the set of visited or closed nodes, and $\mathcal{O}$ the set of unvisited or open nodes, such that $\mathcal{V} = \mathcal{C} \cup \mathcal{O}$. Figure~\ref{fig:BB} illustrates how B\&B operates on an example. Initially,  $GU\!B=\infty$, $\mathcal{O} = \{ v_0\}$, and $\mathcal{C} = \emptyset$.
    At each iteration $t \geq 0$, the node selection policy $\rho : \mathcal{S} \rightarrow \mathcal{O}$ selects the next node to explore. 
    Let $x^*_{LP} \in \mathbb{R}^n$ be the optimal solution to the linear program (LP) relaxation of~$P_{t}$, the MILP associated with the current node $v_t$:
    \begin{itemize} 
        \item If the relaxation of $P_t$ admits no solution,
        $v_t$ is marked as closed and the branch is pruned by infeasibility. If $x^*_{LP} \in \mathbb{R}^n$ exists, and $GU\!B < c^\top x^*_{LP}$, no integer solution in $P_t$ can improve $GU\!B$, thus $v_t$ is marked as closed and the branch is pruned by bound. If $x^*_{LP}$ is not dominated by $\bar{x}$ and $x^*_{LP}$ is feasible, a new incumbent solution $\bar{x} = x^*_{LP}$ has been found.
        Hence, $GU\!B$ is updated and $v_t$ is marked as closed while the branch is pruned by integrity.
        \item Else, $v_t$ is called \textit{branchable}, as $x^*_{LP}$ admits fractional values for some integer variables. The branching heuristic $\pi: \mathcal{S} \rightarrow \mathcal{I}$ selects a variable $x_b$ with fractional value $\hat{x}_b$, to partition the solution space. As a result, two child nodes with associated MILPs~$P_{-} = P_{t} \, \cup \, \{ x_b \leq \lfloor \hat{x}_b \rfloor \}$ and~$P_{_{+}} = P_{t} \, \cup \, \{ x_b \geq \lceil \hat{x}_b \rceil\}$, are added to the set of open nodes $\mathcal{O}$ in place of $v_t$.\footnote{$\hat{x}_b$ denotes the value of $x_b$ in $x^*_{LP}$. We use the symbol $\cup$ to denote the refinement of the bound on $x_b$ in $P_{t}$.} 
    
    \end{itemize}
    
    \greeng{
    This process is repeated until $\mathcal{O} = \emptyset$ and $\bar{x}$ is returned. The dynamics governing the B\&B algorithm between two branching decisions can be described by the function $\kappa_{\rho}: \mathcal{S} \times \mathcal{I} \rightarrow \mathcal{S}$, such that $s' =  \kappa_{\rho}(s, \pi(s))$. By design, B\&B does not terminate before finding an optimal solution and proving its optimality. Consequently, optimizing the performance of B\&B on a distribution of MILP instances is equivalent to minimizing the expected solving time of the algorithm. As ~\citet{etheve_solving_2021} evidenced, the variable selection strategy~$\pi$ is by far the most critical B\&B heuristic in terms of computational performance. In practice, the total number of nodes of the B\&B tree is used as an alternative metric to evaluate the performance of branching heuristics $\pi$, as it is a hardware-independent proxy for computational efficiency. 
    Under these circumstances, given a fixed node selection strategy~$\rho$, the optimal branching strategy~$\pi^*$ associated with a distribution~$p_0$ of MILP instances can be defined as: 
    \begin{equation}
    \label{eq:obj}
        \pi^* = \arg \min_{\pi} \mathbb{E}_{P\sim p_0}(|BB_{(\pi, \rho)}(P)|)
    \end{equation}
    with~$|BB_{(\pi,\rho)}(P)|$ the size of the~$B\&B$ tree after solving $P$ to optimality following strategies ($\pi, \rho$).
    }

    \paragraph{Markov decision process formulation.} The problem of finding an optimal branching strategy according to Eq.~\eqref{eq:obj} can be described as a discrete-time deterministic Markov decision process (MDP) $(\mathcal{S}, \mathcal{A}, \mathcal{T}, p_0, \mathcal{R})$.
    State space $\mathcal{S}$ is the set of all B\&B trees as defined previously, 
    action space $\mathcal{A}$ is the set of all integer variables indices $\mathcal{I}$. 
    Initial states are single node trees, where the root node is associated to a MILP $P_0$ drawn according to the distribution $p_0$. The Markov transition function is defined as~$\mathcal{T} = \kappa_\rho$. Importantly, all states for which $\mathcal{O}=\emptyset$ are terminal states. The reward model is defined as $\mathcal{R}(s,a)=-1$ for all transitions until episode termination. In this setting, the Bellman state-value function
    % $V^\pi: \mathcal{S} \rightarrow \mathbb{R}$
    writes $V^{\pi}(s) = - \mathop{\mathbb{E}}_{a' \sim \pi(s')} [\sum_{t'\geq 0} \gamma^{t'}]$ for $s\in \mathcal{S}$. Since episode horizons are bounded by the (finite) largest possible number of nodes, we take a discount factor $\gamma = 1$. Indeed, under this formulation, the optimal policy $\pi^*$ defined in Eq.~\eqref{eq:obj} also maximizes the state-value function $V^{\pi}(s)$ for all $s \in \mathcal{S}$. Crucially,
    \citet{etheve_reinforcement_2020} and \citet{scavuzzo_learning_2022} have shown that when the B\&B tree is expanded following a depth-first-search (DFS) node selection policy, minimizing the total B\&B tree size is achieved when any subtree is of minimal size. This allows the decomposition of B\&B episodes into independent subtree trajectories, which helps mitigate credit assignment issues arising from the length of B\&B episodes. Moreover, under $\rho = DFS$, $V^\pi$ can be decomposed as the sum of the negative size $\bar{V}^\pi$ of the subtrees rooted in the open nodes of $s \in \mathcal{S}$ : 
    \begin{equation}
        \label{eq:v=w}
        V^{\pi}(s)
        = \sum_{o \in \mathcal{O}} \bar{V}^{\pi}(o, \bar{x}_{o}),
    \end{equation}
    with $\bar{x}_{o} \in \mathbb{R}^{n}$ the incumbent solution when $o$ is selected for expansion. Conveniently, this decomposition enables to derive $V^\pi$ by training graph convolutional neural networks to approximate $\bar{V}^\pi$, using the MILP bipartite graph representation introduced by \citet{gasse_exact_2019} as observation function for B\&B nodes. Thus, as in previous works, we take $\rho = DFS$ in the remainder of the paper.

    \paragraph{MuZero.} Model-based reinforcement learning approaches exploit reversible access to the MDP to design policy improvement operators via planning \cite{silver_general_2018, hansen2023td}. Monte Carlo Tree Search (MCTS) is among the most widely used planning algorithms in MBRL.
    Given access to an environment model $(\mathcal{T}, \mathcal{R})$, MCTS leverages value and policy estimates to select actions leading to promising states, while balancing an exploration-exploitation criterion.
    In order to extend the applicability of MCTS-based RL algorithms to broader control tasks where efficient simulators are not available, \citet{schrittwieser_mastering_2020} introduced MuZero, building upon the previous AlphaZero framework \cite{silver_general_2018}. 
    Concretely, MuZero learns a model consisting of three interconnected networks. First, the representation network $h$ maps raw state observations $s_t$ to a latent state $\hat{s}_t = h(s_t)$. This internal state can then be passed to the prediction network $f$ to obtain state policy and value estimates $( \hat{\mathrm{p}}_t, \hat{\mathrm{v}}_t)$. Alternatively, latent states can be passed along with an action $a_t$ to the dynamics network $g$, to produce predictions for both the true reward $r_t$ and the internal representation $h(s_{t+1})$ of the next visited state when taking action $a_t$ at $s_t$: $g(\hat{s}_t, a_t) = (\hat{r}_t, \hat{s}_{t+1})$.
    At each decision step, MuZero integrates $h$, $f$ and $g$ to perform MCTS, thereby generating improved policy targets $\pi_t$ from which actions are sampled. 
    During training, the model is unrolled for $K$ hypothetical steps and its predictions are aligned with sequences sampled from real trajectories collected by MCTS actors. By enforcing consistency between the predictions observed along simulated rollouts and those observed on real trajectories, MuZero constrains its model to capture only the information most relevant for accurately estimating future states' policy and value. This approach was shown to significantly reduce the burden of MDP dynamics approximation, unlocking the use of MBRL in visually complex domains where model-free RL methods had previously dominated.
    Subsequent work introduced several enhancements to the MuZero framework, including a temporal consistency loss for the dynamics network \citep{ye_mastering_2021} and the use of Gumbel search for improved search efficiency in MCTS \citep{danihelka2022policy}, which proved particularly useful in environments with large action space.

\section{Planning in Branch-and-Bound}
    \label{sec:core}
    In the context of B\&B, taking a step in the environment involves solving the two linear programs associated with the nodes generated by the branching decision. Unfortunately, this procedure is very expensive to simulate, and remains difficult to approximate accurately \citep{qian_exploring_2023}.
    In order to overcome these limitations, we introduce Plan-and-branch-and-bound (PlanB\&B), a model-based reinforcement learning agent adapting the MuZero framework to the B\&B setting. Crucially, our learned model is not explicitly trained to solve linear programs. 
    Instead, it learns the dynamics of an abstract, value-equivalent MDP that retains only the aspects of B\&B essential for enabling policy improvement via MCTS.
    % Instead, it focuses on approximating the dynamics of an abstract, value-equivalent MDP,  capturing only the aspects of B\&B essential for enabling policy improvement via MCTS.}
    
\begin{figure}[t]
        \centering
        \includegraphics[width=0.49\textwidth]{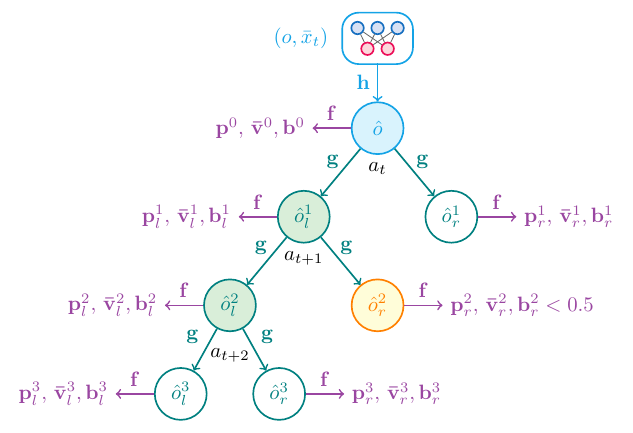}
        \caption{Planning in B\&B over a learned model. The combined use of $h$, $f$, and $g$ allows simulating subtree rollouts starting from the current B\&B node. 
        Here, $\hat{\mathrm{T}}^3 =(\hat{\mathcal{O}}^3, \hat{\mathcal{C}}^3)$ with $\hat{\mathcal{O}}^3 = \{ \hat{o}^3_l, \hat{o}^3_r, \hat{o}^1_r\}$ and $\hat{\mathcal{C}}^3=\{ \hat{o}^2_r\} $.
        To simplify notations, we write $\mathrm{z}^i_j$ in place of $\mathrm{z}_{\hat{o}^i_j}$ for $\mathrm{z} \in \{\mathrm{p}, \bar{\mathrm{v}}, \mathrm{b}\}$.}
        \label{fig:MuZero}
    \end{figure}

    \paragraph{Model.}
    Let $s_t = (\mathcal{V}_t, \mathcal{E}_t, \bar{x}_t) \in \mathcal{S}$ be a B\&B tree where $\mathcal{V}_t = \mathcal{O}_t \cup \mathcal{C}_t$, and let $o = \rho(s_t) \in \mathcal{O}_t$ be the node currently selected for expansion. MuZero is originally designed to operate on full state observations $s_t$ as input to its internal model. In contrast, prior RL and IL branching agents have been designed to rely solely on information from the current B\&B node, represented by the pair $(o, \bar{x}_{t})$, and encoded using the MILP bipartite graph representation function from \citet{gasse_exact_2019}. While, under DFS, this local information is sufficient to recover the optimal policy at state $s_t$, it is generally insufficient to infer the subsequent state $s_{t+1}$. In fact, whenever taking action $a_t$ at $s_t$ leads to fathoming the subtree under $o$, the next visited node will be a leaf node that can not be deduced directly from the triplet $(o, \bar{x}_{t}, a_t)$.   
    To address this, our model learns to predict recursively, \textbf{along subtree trajectories}, the internal representations $(\hat{o}_{l}, \hat{o}_{r})$ associated with the left and right child nodes $(o_{l}, o_{r})$ generated by the agent's branching decisions.\footnote{Following our convention, in DFS, $o_{l}$ designates the node selected immediately next for expansion, while $o_{r}$ designates the node selected once the subtree rooted in $o_{l }$ has been fathomed.}
    Doing so, PlanB\&B enables simulating imagined subtree trajectories while relying exclusively on the information available at the current node. 
    The representation network $h$ first maps the pair $(o, \bar{x}_{t})$ to an internal representation $\hat{o}$ which serves as the root node for the imagined tree $\hat{\mathrm{T}}=(\hat{\mathcal{O}}, \hat{\mathcal{C}})$. Initially, $\hat{\mathcal{O}} = \{\hat{o}\}$ and $\hat{\mathcal{C}} = \emptyset $. The current imagined node $\hat{o}$ can then be passed to the dynamics network $g$ along with any action in $\mathcal{A}$ to generate $(\hat{o}_{l}, \hat{o}_{r})$. Since $r_t = -1$ is constant, $g$ is not tasked with predicting the future reward. However, note that if either of the real child nodes is pruned, either by bound, integrity or infeasibility, its associated subtree value is null, and, consequently, its node internal representation should not be considered for expansion by the model. To distinguish nodes leading to branching decisions from nodes destined to be pruned, we task the prediction network $f$ with estimating future nodes' \textit{branchability} $\mathrm{b} \in \{0,1\}$, in addition to policy and subtree value estimates\footnote{By construction, $\kappa_\rho$ ensures that the current node at $s_t$ is always branchable.}. Based on the predicted branchability values, PlanB\&B updates the sets $(\hat{\mathcal{O}}, \hat{\mathcal{C}})$ by discarding unbranchable nodes and designates the next visited node as the node in $\hat{\mathcal{O}}$ with the greatest depth. If $\hat{\mathcal{O}} = \emptyset$, the imagined subtree has been fully explored, and all subsequent recursive calls to $g$ will, by convention, receive a null reward. The overall interaction between the networks $h$, $f$, and $g$ during the simulation of $k$-step subtree trajectories is illustrated in Figure~\ref{fig:MuZero}. In-depth model description as well as network architectures for $h$, $f$ and $g$ are presented in Appendix B. % \ref{app:features}.

    \paragraph{Data generation.}
    Since $\rho = DFS$, learning a policy minimizing the local subtree size is equivalent to learning a globally optimal policy. Therefore, we can use the model introduced in the previous paragraph to derive improved branching policy targets $\pi_t$ through planning. Due to the large action space size typically encountered in MILP environments, we implement Gumbel search \citep{danihelka2022policy}, a variant of MCTS presented in Appendix C. %\ref{app:mcts}. 
    % \greenv{In fact, Gumbel search uses the Gumbel Top-$k$ trick \cite{kool2019stochastic} at the root of the search tree to reduce simulation costs while preserving search efficiency.}
    Throughout the search, the policy prior $\hat{\mathrm{p}}^k_t$ associated with the imagined tree $\hat{\mathrm{T}}^{k}$ is given by the policy prediction of the branchable node in $\hat{\mathcal{O}}^k$ with the greatest depth. 
    Similarly, following Eq.~\eqref{eq:v=w}, the estimate state value $\hat{\mathrm{v}}^k_{t}$ is obtained by summing the subtree value predictions $\bar{\mathrm{v}}_{\hat{o}}$ of all open nodes in the imagined tree: $\hat{\mathrm{v}}^k_{t} = \sum_{\hat{o} \in \hat{\mathcal{O}}^{k}} \bar{\mathrm{v}}_{\hat{o}}$.  
    Thus, by combining networks $h$, $f$ and $g$ to simulate subtree rollouts, Gumbel search yields an improved target policy $\pi_t$ from which the selected action $a_t$ is sampled. Repeating this process generates B\&B episodes that are stored in memory and subsequently used for training.

    \paragraph{Learning.}
     The agent is trained over $K$-step subtree trajectories $(s_t, a_t,\dots, s_{t+K})$ sampled from memory. As in MuZero, the model is unrolled from $s_{t}$ over $K$ steps to generate imagined trajectories $(\hat{\mathrm{T}}^{0},\dots, \hat{\mathrm{T}}^{K})$.
     As illustrated in Figure \ref{fig:MuZero}, for each step $k=0,\dots,K$, the model recursively predicts $\hat{\mathrm{p}}^k_t$ and $\hat{\mathrm{v}}^k_t$ to approximate the Gumbel search policy $\pi_{t+k}$ and $n$-step return $z_{t+k} = -n + \hat{\mathrm{v}}^{k+n}_{t}$ prediction targets. 
     Moreover, the predicted branchability scores of imagined nodes are trained to match the true branchability labels of real nodes.
     To support this, we introduce a new self-supervised loss term $\mathcal{L}_{\mathrm{T}}$, which enforces structural consistency between real and imagined subtrees, as well as hierarchical consistency between consecutive B\&B node observations.
     The overall loss minimization objective can thus be summarized as:
    \begin{equation*}
        \mathcal{L}_t(\theta) = \sum_{k=0}^{K} \mathcal{L}_\mathrm{p}(\pi_{t+k}, \hat{\mathrm{p}}^k_{t}) + \mathcal{L}_\mathrm{v}(z_{t+k}, \hat{\mathrm{v}}^k_{t}) + \mathcal{L}_{\mathrm{T}}(s_{t+K}, \hat{\mathrm{T}}_{K})
    \end{equation*}

    with $\theta$ the global parameter for $f$, $g$, $h$. Losses $\mathcal{L}_{\mathrm{p}}, \mathcal{L}_{\mathrm{v}}$ and $\mathcal{L}_{\mathrm{T}}$ are presented in Appendix E. %\ref{app:loss}.

\section{Related work}
    % \greeng{Following the seminal work by \citet{gasse_exact_2019}, several contributions have proposed to build more complex neural network architectures based on transformers \citep{lin_learning_2022} and recurrence mechanisms \citep{seyfi_exact_2023} to improve the performance of IL branching agents, with limited success. In parallel, theoretical and computational analysis \citep{bestuzheva_scip_2021, sun_improving_2022} have shown that neural networks trained by imitation could not rival the tree size achieved by strong branching (SB), the branching expert used in \citet{gasse_exact_2019}. In fact, low tree sizes associated with SB turn out to be primarily due to the formulation improvements resulting from the massive number of LPs solved in SB, not to the intrinsic quality of the branching decisions themselves.}
    Following the work of \citet{gasse_exact_2019}, several studies have explored adopting more expressive neural architectures, from transformer-based models \citep{lin_learning_2022} to recurrent designs \citep{seyfi_exact_2023}, for the purpose of learning to branch. However, these architectural extensions have yielded only limited gains. In parallel, theoretical and empirical analyses \citep{dey_theoretical_2021} have shown that the small trees produced by strong branching (SB), the branching expert used in \citet{gasse_exact_2019}, arise mainly from the extensive formulation tightening induced by solving large numbers of LPs, rather than from the intrinsic quality of its branching decisions.
    
    Such findings cast doubt on the long-term effectiveness of purely imitation-based approaches.
    Since branching decisions unfold sequentially, reinforcement learning provides a principled alternative framework for learning to branch.
    Building on the works of \citet{etheve_reinforcement_2020} and \citet{scavuzzo_learning_2022}, \citet{parsonson_reinforcement_2022} explored training RL branching agents on B\&B trees expanded following solvers' default node selection policies, rather than DFS.
    To address the partial observability induced by moving away from DFS, they trained their agent on retrospective trajectories, tree-diving trajectories extracted from original B\&B episodes. 
    %More generally, a large body of work has proposed to learn, either by imitation or reinforcement, better-performing B\&B heuristics outside of variable selection \citep{nair_solving_2021, paulus_learning_2022, sonnerat_learning_2022}.
    
    More generally, a substantial line of research \citep{nair_solving_2021, paulus_learning_2022} has explored augmenting B\&B heuristics with machine learning algorithms beyond variable selection.
    Building on the MDP formulation first proposed by \citet{he_learning_2014}, recent RL contributions to primal search \citep{sonnerat_learning_2022, wu_deep_2023}, node selection \citep{etheve_solving_2021, zhang_learning_2024}, and cut selection \citep{tang_reinforcement_2020, wang_learning_2023} have primarily focused on deploying traditional model-free RL algorithms within the B\&B framework, adapting the action space to suit the specific heuristic being targeted. 
    
    %Finally, machine learning applications to CO are not limited to B\&B. For example, in the context of routing or scheduling problems, where exact resolution rapidly becomes prohibitive, agents are trained to learn direct search heuristics yielding high-quality feasible solutions \citep{kool_attention_2019, chalumeau2023combinatorial}.
    Finally, machine learning methods have been applied beyond the scope of exact combinatorial optimization. In large routing and scheduling problems, where exact resolution quickly becomes intractable, recent works have trained neural agents to act as direct-search heuristics, capable of producing strong feasible solutions \citep{kool_attention_2019, chalumeau2023combinatorial}.
    Drawing inspiration from model-based planning, \citet{pirnay2024self} proposed a self-improvement operator that refines learning agents' policies by generating alternative trajectories via stochastic beam search, and treating improved trajectories as targets for imitation learning. Applied to routing problems, their approach matched the performance of imitation learning agents trained on expert demonstrations, highlighting the potential of planning-based methods as a data-efficient alternative to expert supervision in combinatorial optimization.

\section{Experimental study}
\label{sec:exp}

     We now assess the efficiency of our model-based branching agent, as we aim to answer the following questions:
     % We now conduct an experimental study to assess the efficiency of our model-based RL branching agent. In particular, we aim to answer the following questions.
    \begin{itemize}[label={},  left=0pt, labelsep=0.0em, itemindent=0pt, align=left]
        \item[\textbf{Q1}] \hspace{1mm} Can PlanB\&B learn an efficient policy network to guide MILP solving? In particular, how does it compare against solver heuristics, as well as prior RL and IL approaches?
        \item[\textbf{Q2}] \hspace{1mm} When provided with additional search budget, can our agent further improve the quality of its decisions by leveraging its internal model of B\&B?
        \item[\textbf{Q3}] \hspace{1mm} To what extent does the branching behavior of PlanB\&B align with that of the expert strong branching (SB) strategy?
        \item[\textbf{Q4}] \hspace{1mm} Is the use of a DFS node selection policy inherently detrimental to the performance of branching agents?
    \end{itemize}
    \begin{table*}[t]
        \centering
        \small
        \begin{tabular}{ccccccccccccccc} 
             & \multicolumn{2}{c}{\textbf{Set Covering}} & \multicolumn{2}{c}{\textbf{Comb. Auction}} & \multicolumn{2}{c}{\textbf{Max. Ind. Set}} & \multicolumn{2}{c}{\textbf{Mult. Knapsack}} & \multicolumn{4}{c}{\textbf{Norm. Score $\mathbf{\pm}$ Conf. Interval}} \\ 
            Method & Node & Time & Node & Time & Node & Time & Node & Time & Node & CI & Time & CI \\
            % \toprule
            \cmidrule(r){1-9} \cmidrule(l){10-13}
            Presolve & $-$ & $4.74$ & $-$  & $0.90 $ & $-$  & $1.78 $ & $-$  & $0.20$ & $-$ & $-$ & $-$ & $-$\\
            % \midrule
            \cmidrule(r){1-9} \cmidrule(l){10-13}
            Random & $3289$ & $5.94$ & $1111$ & $2.16$ & $386.8$ & $2.01$ & $733.5$ & $0.55$ & $1068$ & $\pm 113 $& $454$ & $\pm 51$\\
            SB & $35.8 $ & $12.93$ & $28.2$ & $6.21 $ & $24.9$ & $45.87$ & $161.7$ & $0.69$& $46$ & $\pm 0$ & $4279$ & $\pm 57$\\
            SCIP & $62.0 $ & $2.27 $ & $20.2 $ & $1.77$ & $19.5$ & $2.44 $ & $289.5 $ & $0.53$ & $58$ & $\pm 0$&  $\greent{363}$ & $\pm 3$\\
            \cmidrule(r){1-9} \cmidrule(l){10-13}
             ${\text{IL}^\star}$ & \textcolor{purple}{$\mathbf{133.8}$}  & $0.90 $ & \textcolor{purple}{$\mathbf{83.6}$} & $0.65$ & \textcolor{purple}{$\mathbf{40.1}$} & $0.36$ & {$272.0$} &{$0.69$} & \textcolor{purple}{$\mathbf{96}$} & \textcolor{purple}{$\mathbf{\pm17}$} & \greent{$116$} & $\pm 15$\\
             IL-DFS & $136.4$ & \textcolor{purple}{$\mathbf{0.74}$} & $92.1$ & $0.56$ & $68.5$ & $0.44$ & $411.5$ & $1.07$ & $\greent{130}$ & $\pm26$& \greent{$131$} & $\pm 20$\\
            \cmidrule(r){1-9} \cmidrule(l){10-13}
            PG-tMDP & $649.4 $ & $2.32 $ & $168.0 $ & $0.94 $ & $153.6 $ & $0.92$ & $436.9$ & $1.57$ & \greent{$272$} & $\pm 46$ & \greent{$254$} & $\pm 50 $ \\
            DQN-tMDP & \textcolor{blue}{$\mathbf{175.8}$} & \textcolor{blue}{$\mathbf{0.83}$} & $203.3$ & $1.11$ & $168.0$ & $1.00$ & $266.4$ & $0.73$ & \greent{$207$} & $\pm 25$& \greent{$188$} & $\pm 16$\\
            DQN-Retro & $183.0 $ & $1.14 $ & $103.2$ & $0.78$ & $223.0$ & $1.81$ & $250.3$ & $0.67$ & \greent{$208$} & $\pm 23$ & \greent{$241$} & $\pm 25$\\
            PlanB\&B & 186.2 & 0.87 & 
            \textcolor{blue}{$\mathbf{84.7}$} & \textcolor{purple}{$\mathbf{0.54}$} & \textcolor{blue}{$\mathbf{44.8}$} & \textcolor{purple}{$\mathbf{0.32}$} & \textcolor{purple}{$\mathbf{220.0}$} & \textcolor{purple}{$\mathbf{0.55}$} & \textcolor{blue}{$\mathbf{100}$} & \textcolor{blue}{$\mathbf{\pm9}$} & \textcolor{purple}{$\mathbf{100}$} & \textcolor{purple}{$\mathbf{\pm 12}$} \\
            % \bottomrule[1.5pt]
            \cmidrule[1.5pt](r){1-9} \cmidrule[1.5pt](l){10-13}
            \multicolumn{13}{c}{Test instances} \\
        \hspace{5mm}
        \end{tabular}
        \begin{tabular}{ccccccccccccccc} 
             & \multicolumn{2}{c}{\textbf{Set Covering}} & \multicolumn{2}{c}{\textbf{Comb. Auction}} & \multicolumn{2}{c}{\textbf{Max. Ind. Set}} & \multicolumn{2}{c}{\textbf{Mult. Knapsack}} & \multicolumn{4}{c}{\textbf{Norm. Score $\mathbf{\pm}$ Conf. Interval}} \\ 
            Method & Node & Time & Node & Time & Node & Time & Node & Time & Node & CI & Time & CI \\
            \cmidrule(r){1-9} \cmidrule(l){10-13}
            Presolve & $-$ & $\greent{12.3}$ & $-$ & $\greent{2.67}$ & $-$ & $\greent{5.16}$ & $-$ & $\greent{0.46}$& $-$ & $-$ & $-$ & $-$ \\
           \cmidrule(r){1-9} \cmidrule(l){10-13}
            Random & $271632$  & $842$ &  $317235$  & $749$ & $215879$ & $2102$ & $93452$ & $70.6$& \greent{$8050$} & $\pm 1646$ & \greent{$3870$} & $\pm 847$\\
            SB & $672.1$ & $398$ & $389.6$  & $255$ & $169.9$ & $2172$ & \textcolor{purple}{$\mathbf{1709}$} & \textcolor{purple}{$\mathbf{12.5}$} & \greent{$13$} & $\pm 0$ & \greent{$2243$} & $\pm 47$\\
            SCIP & $3309$  & $48.4$ &  $1376$  & $14.77$ & $3368$ & $90.0$ & $30620$ & $22.1$ & \greent{$121$} & $\pm 0$& \greent{$132$} & $\pm 1$\\
            \cmidrule(r){1-9} \cmidrule(l){10-13}
             ${\text{IL}^\star}$ & \textcolor{purple}{$\mathbf{2610}$}  & {$23.1$} &  \textcolor{purple}{$\mathbf{1282}$}  & 9.4 & \textcolor{purple}{$\mathbf{1993.0}$} & \textcolor{purple}{$\mathbf{38.6}$} & $11730$ & $43.5$ & \textcolor{purple}{$\mathbf{70}$} & \textcolor{purple}{$\mathbf{\pm7}$}& \textcolor{purple}{$\mathbf{83}$} & \textcolor{purple}{$\mathbf{\pm7}$}\\
             IL-DFS & $3103 $ & \textcolor{purple}{$\mathbf{22.5}$} & $1828 $ & $10.2 $ & $3348$ & $51.9$& $43705$ & $130.8$ & \greent{$151$} & $\pm 30$ & \greent{$136$} & $\pm 21$\\
            \cmidrule(r){1-9} \cmidrule(l){10-13}
            PG-tMDP & $44649 $ & $221 $ & $6001 $ & $30.7 $ & $3133$ & $43.6$ & $35614$ & $123$ & \greent{$373$} &  $\pm48$ & \greent{$290$} & $\pm42$\\
            DQN-tMDP & $8632$  & $71.3$ &  $20553$  & $116$ & $45634$ & $477$ & $22631$ & $65.1$ & \greent{$787$} & $\pm 128$& \greent{$679$} & $\pm 78$\\
            DQN-Retro & $6100$  & $59.4$ &  $2908$  & $18.4$ & $119478$ & $1863$ & $27077$ & {$79.5$} & \greent{$1166$} & $\pm256$ & \greent{$1254$} & $\pm 218$\\
            PlanB\&B & \textcolor{blue}{$\mathbf{5869}$}  & \textcolor{blue}{$\mathbf{46.2}$} & \textcolor{blue}{$\mathbf{1665}$} & \textcolor{purple}{$\mathbf{9.1}$} & \textcolor{blue}{$\mathbf{2853}$} & \textcolor{blue}{$\mathbf{41.1}$} & \textcolor{blue}{$\mathbf{13574}$} & \textcolor{blue}{$\mathbf{51.2}$} & \textcolor{blue}{$\mathbf{100}$} & \textcolor{blue}{$\mathbf{\pm9}$} &\textcolor{blue}{$\mathbf{100}$} & \textcolor{blue}{$\mathbf{\pm 8}$}\\
            % \bottomrule[1.5pt]
            \cmidrule[1.5pt](r){1-9} \cmidrule[1.5pt](l){10-13}
            \multicolumn{13}{c}{Transfer instances} \\
        \end{tabular}
            \caption{Performance comparison of branching agents on four standard MILP benchmarks. For each method, we report the total number of B\&B nodes, presolve time, and total solving time excluding presolve. The presolve phase is identical across all methods. Lower values indicate better performance. \textcolor{purple}{\textbf{Red}} highlights the overall best agent, while \textcolor{blue}{\textbf{blue}} marks the best-performing RL-based agent. Following prior work, results are reported as the geometric mean over 100 unseen test instances and an additional \greent{100} higher-dimensional transfer instances, averaged across 5 random seeds. \greent{Norm. Score represents the aggregate average performance of each agent across the four MILP benchmarks, normalized by the score of PlanB\&B. Aggregate confidence intervals (CI) are reported for each baselines; additional per-benchmark confidence intervals are provided in Appendix~I.}}
        \label{tab:results}
    \end{table*}
    \label{sec:exp_setup}
    \paragraph{Benchmarks.} We consider four standard MILP benchmarks: set covering (SC), combinatorial auctions (CA), maximum independent set (MIS) and multiple knapsack (MK) problems. 
    In our experiments, SCIP 8.0.3 \citep{bestuzheva_scip_2021} is used as backend MILP solver, along with the Ecole library \citep{prouvost_ecole_2020} for instance generation, see Appendix A for further benchmark details. 
    \paragraph{Baselines.}  We compare our PlanB\&B agent against prior RL agents, namely DQN-tMDP \citep{etheve_reinforcement_2020}, PG-tMDP \citep{scavuzzo_learning_2022} and DQN-Retro \citep{parsonson_reinforcement_2022}. 
    We also compare against the IL expert from \citet{gasse_exact_2019}, evaluated under both SCIP’s default node selection policy (${\text{IL}^\star}$), and DFS (IL-DFS). More details on these baselines can be found in Appendix G. %\ref{app:baseline}.
    Finally, we report the performance of reliability pseudo cost branching (SCIP), the default branching heuristic used in SCIP, strong branching (SB) \citep{applegate_finding_1995}, the greedy expert from which the IL agent learns from, and random branching (Random), which randomly selects a fractional variable.
    SCIP configuration is common to all baselines. 
    As in prior works, we set the time limit to one hour, disable restart, and deactivate cut generation beyond root node. 
    All the other parameters are left at their default value.
    \label{sec:train}
    \paragraph{Training \& evaluation.} The overall PlanB\&B training pipeline is provided in Appendix F. %\ref{app:rl}.
    Branching agents are trained on instances of each benchmark separately.
    For evaluation, we report performance in terms of both node count and solving time on 100 test instances unseen during training, as well as on 100 transfer instances of higher dimensions. 
    Evaluation metrics are averaged over 5 random seeds.
    Importantly, when comparing ML-based branching strategies to standard SCIP heuristics such as RPB or SB, solving time remains the only reliable performance indicator. This is because invoking a custom branching rule in SCIP triggers auxiliary routines, such as conflict analysis and bound tightening, that strengthen the MILP formulation while incurring computational overhead.
    \paragraph{Baselines comparison (Q1)} 
    Computational results obtained on the four benchmarks are presented in Table \ref{tab:results}. Standard deviations, additional performance metrics as well as targeted ablations are provided in Appendix I.
     Unlike conventional MBRL settings, PlanB\&B operates under strict time and computational constraints.
     Crucially, every second spent on planning directly increases the overall solving time, thereby impacting final performance.
     Therefore, to allow systematic comparison, the results reported for PlanB\&B in Table \ref{tab:results} reflect the performance of its policy network alone, without any computational budget allocated to MCTS simulations at evaluation time.
     On aggregate test instances, compared to prior RL baselines, PlanB\&B’s policy network achieves $2\times$ reductions both in tree size and solving time.
     This performance gap broadens further to $3\times$ on aggregate transfer instances, underscoring the superior generalization capabilities of PlanB\&B relative to prior RL agents. 
     PlanB\&B also outperforms the IL-DFS agent on both test and transfer instances, providing, to our knowledge, the first evidence of an RL-based branching strategy surpassing an IL agent trained to mimic strong branching. Specifically, our experiments demonstrate that, under a DFS node selection policy, PlanB\&B learns branching strategies superior to that of the IL agent from \citet{gasse_exact_2019}. When compared against the standard ${\text{IL}^\star}$ baseline, PlanB\&B manages to achieve $\approx10\%$ lower solving time on test instances, despite producing $\approx5\%$ larger trees in average. However, on transfer instances, PlanB\&B is globally unable to overcome the performance limitations imposed by DFS, although it still manages to outperform the ${\text{IL}^\star}$ baseline on combinatorial auction. 
     The influence of DFS on the performance of branching agents is further analyzed in \textbf{(Q4)}.
     Finally, PlanB\&B clearly outperforms the default SCIP baseline on both test and transfer instances, despite operating under DFS, a performance achieved by no other learning-based baseline.
    
\begin{figure}[t]
        \centering
        \begin{subfigure}{0.4\textwidth}
            \centering
            \includegraphics[width=\linewidth]{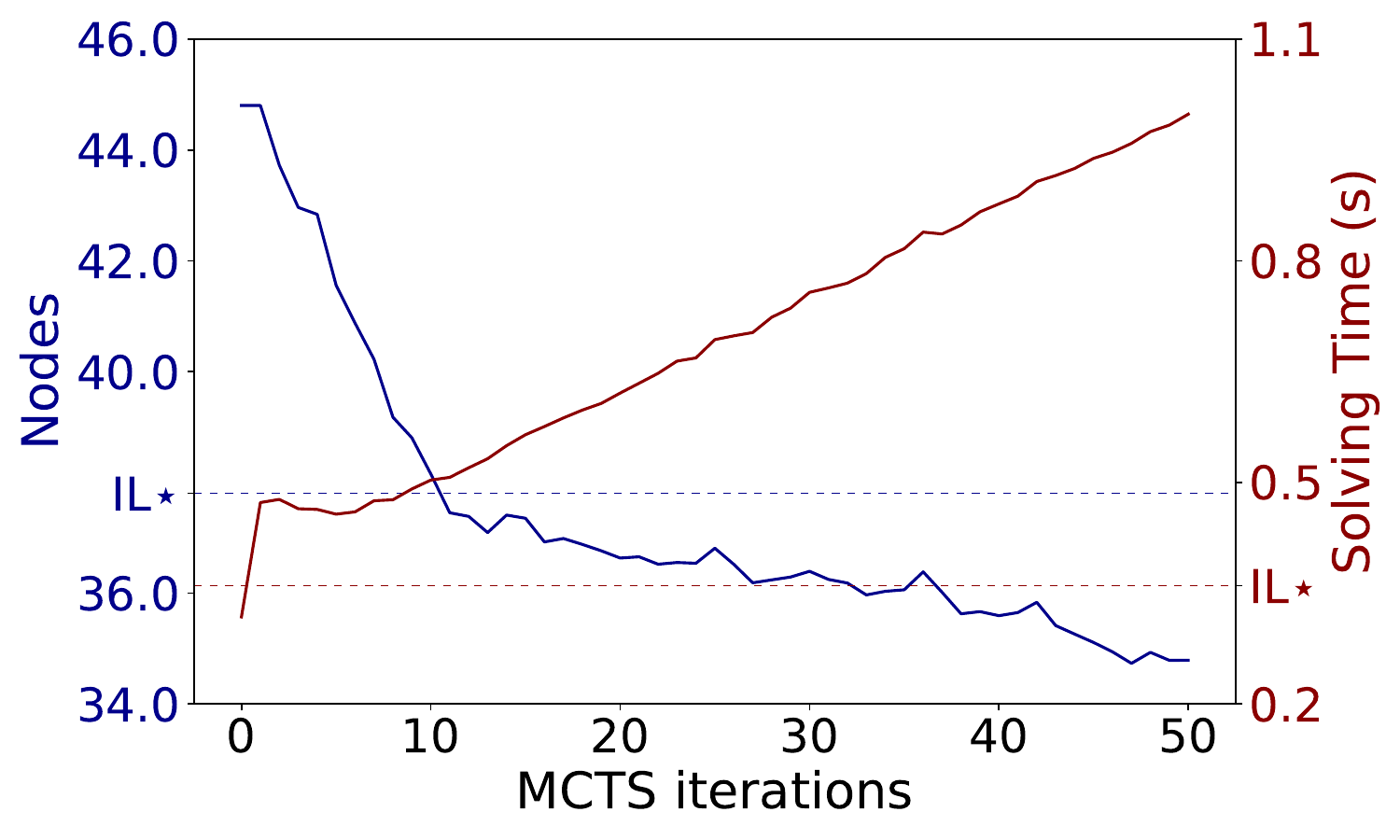}
            \caption{Test instances}
            \label{fig:chart1}
        \end{subfigure}
        %\vspace{1em}  % Adds vertical spacing between the subfigures
        \begin{subfigure}{0.4\textwidth}
            \centering
            \includegraphics[width=\linewidth]{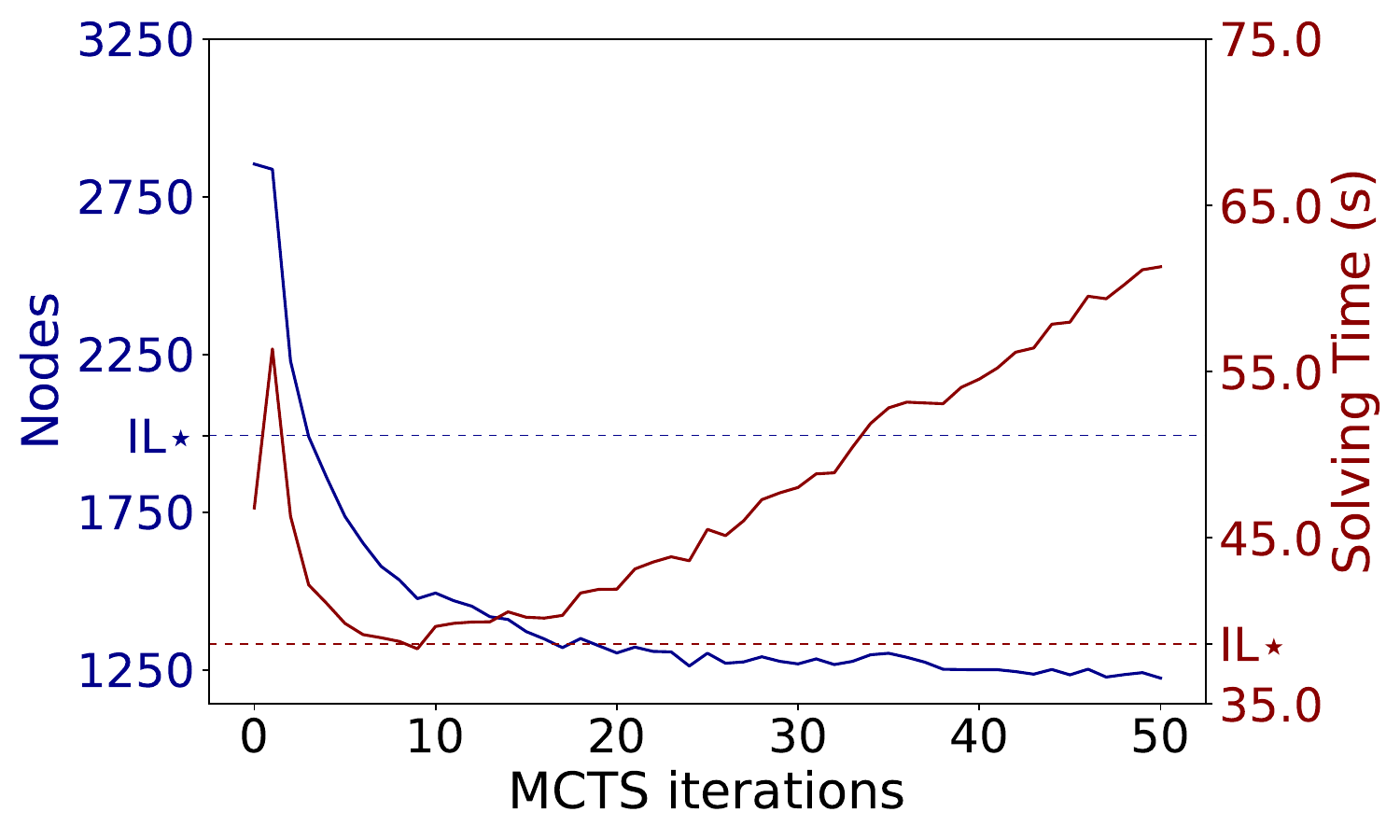}
            \caption{Transfer instances}
            \label{fig:chart2}
        \end{subfigure}
    
        \caption{Policy improvement associated with increased simulation budget over the MIS benchmark.}
        \label{fig:mcts_combined}
    \end{figure}

     \paragraph{Planning in B\&B (Q2)} Although PlanB\&B demonstrates strong performance while relying solely on its policy network, we further investigate its capacity to derive stronger policies by leveraging its learned model to plan at evaluation time. Figure \ref{fig:mcts_combined} shows the effect of increasing PlanB\&B's simulation budget $N$ on both node and time performance over MIS test and transfer instances. On test instances, PlanB\&B achieves lower average tree size than the ${\text{IL}^\star}$ baseline as soon as $N > 12$, reaching up to $\approx20\%$ tree size reduction for $N = 50$. On transfer instances, not only increased simulation budget enables to reduce average tree size up to $\approx50\%$, but this reduction also translates into improved solving time performance, with best solving time, achieved at $N = 9$, matching the solving time performance of the ${\text{IL}^\star}$ agent. Remarkably, when planning over its internal model, PlanB\&B produces branching strategy yielding smaller trees than that produced by the ${\text{IL}^\star}$ agent, despite operating under DFS. 
   
    \paragraph{Is PlanB\&B learning to strong branch? (Q3)} Given the performance trends observed in Figure \ref{fig:mcts_combined}, a natural question arises: does PlanB\&B manage to derive better branching decisions than IL baselines by following SB more closely, or does it discover genuinely novel strategies? 
    After all,
    %as discussed in Appendix D,
    strong branching can be interpreted as a one-step MCTS procedure aimed at maximizing immediate dual gap reduction. To answer this, Table~\ref{tab:SB scores} reports several alignment metrics designed to assess which of the ${\text{IL}^\star}$ and PlanB\&B policies more closely resembles strong branching.
    Across all metrics, PlanB\&B policies exhibit lower alignment with SB than ${\text{IL}^\star}$ policies. 
    Remarkably, the refined branching policy returned by the MCTS is roughly as close to SB as the policy yielded by the policy network. 
    This suggests that PlanB\&B surpasses IL baselines not through closer imitation of strong branching, but by discovering and exploiting original strategies.
    
    \begin{table}[t]
        \centering
        \small
        \begin{tabular}{cccccc}
            &  &  & ($\downarrow$) SB & ($\uparrow$) SB & ($\uparrow$) SB\\
         Instances & Policy & Iter. & C-Entropy & Score & Freq. \\
        \toprule

            & SB & $-$ &  $0.00$ & $1.00$ & $1.00$ \\
        \midrule
        \multirow{3}{*}{Test}
            & ${\text{IL}^\star}$ & $-$ & $0.84$ & $0.69$ & $0.45$ \\
            & PlanB\&B     & $0$ & $1.97$ & $0.63$ & $0.39$ \\
            & PlanB\&B   & $50$ & $2.08$ & $0.65$ & $0.40$ \\
            % & SB & $-$ &  $0.00$ & $1.00$ & $1.00$ \\
            % & Random       & $1.0$ & $0.0$ & $0.002$ \\
        \midrule
        \multirow{3}{*}{Transfer} 
            %& SB & $-$& $0.00$ & $1.00$ & $1.00$ \\
            & ${\text{IL}^\star}$        & $-$   & $0.86$ & $0.76$ & $0.40$ \\
            & PlanB\&B   & $0$  & $1.46$ & $0.72$ & $0.39$\\
            & PlanB\&B   & $50$ & $1.34$ & $0.71$ & $0.38$ \\
            % & Random       & $1.0$ &  $0.0$ & $0.001$ \\
        \bottomrule
        \end{tabular}
        \caption{Alignment metrics between ML baselines and SB on MIS test and transfer instances. Obviously, SB is perfectly aligned with SB. Experimental setup and thorough metrics description is provided in Appendix I.}
        \label{tab:SB scores}
    \end{table}

    \paragraph{Influence of DFS (Q4)}  We finally turn to analyzing the impact of DFS on the performance of branching agents. From Table~\ref{tab:results}, two key observations emerge. First, the performance gap between DFS and non-DFS variants varies significantly across benchmarks: for instance, MK exhibits a particularly large gap, while SC shows a much narrower one. Second, across all benchmarks, this gap consistently widens as the problem dimensionality increases from test to transfer instances. This trend can be attributed to the typical \textit{discovery step} $t_d$ at which the B\&B algorithm discovers the global optimal solution $x^*$. In fact, as soon as $\bar{x} = x^*$, the primal gap is closed, and, consequently, all node selection policies are equivalent for $t \geq t_d$. Therefore, in theory, the smaller the value of $t_d$, the smaller the performance gap. Table~\ref{tab:DFS} reports both absolute and relative average discovery times obtained by IL on SC and MK benchmarks.
    The results are consistent with our theoretical intuition, see Appendix I for further discussion. Moreover, they highlight a fundamental limitation of applying DFS when solving higher-dimensional MILPs: as problem size grows, closing the primal gap becomes increasingly difficult, which in turn exacerbates the computational burden associated with DFS.
    To address this issue, future RL contributions need moving beyond the traditional MILP bipartite graph representation and adopt more expressive global representations of the B\&B tree, as recently proposed by \citet{zhang_learning_2024}.
    
    \begin{table}[t]
        \centering
        \small
        \begin{tabular}{ccccc}
        & SC Test & SC Transfer & MK Test & MK Transfer \\
        \toprule
            $t_d$ & $29$ & $514$ & $376 $ & $10934 $ \\
            $t_r$ & $0.14 $ & $0.25$ & $1.0 $ & $1.0 $ \\
        \bottomrule
        \end{tabular}
        \caption{${\text{IL}^\star}$ agent discovery times, with $t_r = t_d / T$ on SC and MIS benchmarks, see Appendix I for further results.}
        \label{tab:DFS}
    \end{table}
    
\section{Conclusion and perspectives}
Combinatorial optimization has proven in the past to be a challenging setting for traditional RL approaches.
In this work, we introduced PlanB\&B, a novel model-based reinforcement learning framework leveraging a learned internal model of B\&B to discover new variable selection strategies. 
Our experimental study leads to three main findings.  
First, PlanB\&B's dynamics network approximates LP resolution in the latent space with sufficient fidelity to enable policy improvement through model-based planning, thereby extending the applicability of MBRL algorithms originally developed for combinatorial board games to mixed-integer linear programming. 
Second, in the context of repeated MILPs, our MBRL agent learns branching strategies that clearly outperform both SCIP and former RL baselines. Moreover, further analysis shows instances where PlanB\&B surpasses IL, not by more closely replicating expert behavior, but by actively diverging from strong branching patterns, highlighting the potential of RL to uncover branching strategies going beyond existing expert heuristics.
Third and finally, our computational study highlights the burden imposed by the DFS node selection policy when seeking to solve higher-dimensional MILPs. In order to fully unlock the potential of MBRL in exact combinatorial optimization, we expect future research to explore the design of scalable observation functions capable of efficiently encoding evolving B\&B trees.

\bibliography{ref}
\newpage
\section*{A \; Instance dataset}
\label{app:milps}
    Instance datasets used for training and evaluation are described in Table \ref{tab:milps}. We trained and tested on instances of same dimensions as \citet{scavuzzo_learning_2022} and \citet{parsonson_reinforcement_2022}. As a reminder, the size of action set  $\mathcal{A}$ is equal to the number of integer variables in $P$. Consequently, action set sizes in the Ecole benchmark range from 30 to 480 for train / test instances and from 50 to 980 for transfer instances. \\
    \begin{table*}[t]
        \centering
        \footnotesize
        \begin{tabular}{ccccccc} 
            & & & \multicolumn{2}{c}{Parameter value} & \multicolumn{2}{c}{\# Int. variables} \\
            Benchmark & Generation method & Parameters & Train / Test  & Transfer & Train / Test & Transfer \\ 
            \toprule
            \begin{tabular}{@{}c@{}}Combinatorial \\ auction \end{tabular} & \citet{leyton2000towards} & \begin{tabular}{@{}c@{}}Items \\ Bids\end{tabular} & \begin{tabular}{@{}c@{}}100 \\ 500\end{tabular} & \begin{tabular}{@{}c@{}}200 \\ 1000\end{tabular} & $ 100 $ & $200$\\
            \midrule
            Set covering & \citet{balas1980set} &\begin{tabular}{@{}c@{}}Items \\ Sets\end{tabular}& \begin{tabular}{@{}c@{}}500 \\ 1000\end{tabular} & \begin{tabular}{@{}c@{}}1000 \\ 1000\end{tabular} & $ 100$ & $ 130$\\
            \midrule
            \begin{tabular}{@{}c@{}}Maximum \\ independent set\end{tabular}  & \citet{bergman2016decision} & Nodes & 500 & 1000 & $480$ & $ 980$ \\
            \midrule
            \begin{tabular}{@{}c@{}}Multiple \\ knapsack\end{tabular} & \citet{fukunaga2011branch} & \begin{tabular}{@{}c@{}}Items\\ Knapsacks\end{tabular} & \begin{tabular}{@{}c@{}}100 \\ 6\end{tabular} & \begin{tabular}{@{}c@{}}100 \\ 12\end{tabular} & $ 30 $ & $ 50$ \\
            \bottomrule
        \hspace{5mm}
        \end{tabular}
        \caption{Instance size for each benchmark. Performance is evaluated on test instances that match the size of the training instances, as well as on larger instances, to further assess the generalization capacity of our agents. Last two columns indicate the approximate number of integer variables after presolve, both for train / test and transfer instances.}
        \label{tab:milps}
    \end{table*}

\section*{B \; Model architecture}
\label{app:features}
In this section, we formally describe the overall architecture and prediction mechanism of the model introduced in Section 3. 
\paragraph{State representation}
Following the works of \citet{gasse_exact_2019}, MILPs are best represented by bipartite graphs $\mathcal{G} = (\mathcal{V}_{\mathcal{G}}, 
\mathcal{C}_{\mathcal{G}}, \mathcal{E}_{\mathcal{G}})$ where 
$\mathcal{V}_{\mathcal{G}}$ denotes the set of variable nodes, 
$\mathcal{C}_{\mathcal{G}}$ denotes the set of constraint nodes, and 
$\mathcal{E}_{\mathcal{G}}$ denotes the set of edges linking variable and 
constraints nodes.
Nodes $v_{\mathcal{G}} \in \mathcal{V}_{\mathcal{G}}$ and $c_{\mathcal{G}} \in \mathcal{C}_{\mathcal{G}}$ are connected through the edge $\epsilon_{\mathcal{G}} \in \mathcal{E}_{\mathcal{G}}$ if and only if the variable associated with $v_{\mathcal{G}}$ appears in the constraint associated with $c_{\mathcal{G}}$.
% Given a MILP $P$, defined as in Section \ref{sec:b&b}, its associate bipartite representation $\mathcal{G}$ has $|\mathcal{G}| = |\mathcal{V}_{\mathcal{G}}| + |\mathcal{C}_{\mathcal{G}}| = n + m$ nodes.
 We note respectively $d_v$, $d_c$, $d_e$ the input dimension of variable nodes, constraint nodes and edges. Bipartite graphs can be thus interpreted as vectors from the observation space $\mathcal{M} = \mathbb{R}^{n\times d_v} \times \mathbb{R}^{m\times d_c} \times \mathbb{R}^{n \times m \times d_e}$.
In our experiments, IL and PG-tMDP agents use the list of features of \citet{gasse_exact_2019} to represent variable nodes, constraint nodes and edges, while PlanB\&B, DQN-TreeMDP and DQN-Retro agents also make use of the additional features introduced by \citet{parsonson_reinforcement_2022}.

\paragraph{Representation network}
The representation network $h : \mathcal{M} \rightarrow \mathbb{R}^{n\times d_{h}}\times\mathbb{R}^{m\times {d_{h}}}\times\mathbb{R}^{n\times m \times d_{h}}$, maps graph bipartite observations to a triplet of embeddings vectors of higher dimension $d_{h}$.
It is composed of three fully-connected modules: the variable encoder $\mathbf{e}_v$, the constraint encoder $\mathbf{e}_c$ and the edge encoder $\mathbf{e}_e$, which independently compute initial embeddings for the corresponding components.
% followed by two graph convolutionnal message passing operations $\mathbf{\mathcal{G}}_{c \rightarrow v}$ identical to the one introduced by \citet{gasse_exact_2019}.
In the following, we note $\mathcal{H} = \mathbb{R}^{n\times d_{h}}\times\mathbb{R}^{m\times d_{h}}\times\mathbb{R}^{n\times m \times d_{h}}$ the latent space associated with internal representations.

\subsubsection*{Prediction network}
% The branchability of a node equals $1$ if the node is branchable, and $0$ otherwise.
The prediction network $f$ consists of a shared convolutional core followed by three output heads $\mathbf{p}, \bar{\mathbf{v}}, \,\mathbf{b}$, which maps internal representations $\hat{o} \in \mathcal{H}$ to their estimated policy, subtree value and branchability score respectively. In order to enhance its generalization capacity to higher-dimensionnal MILP instances with higher associated tree size, the value network represents subtree values as histogram distributions over the support of the value function. 
% Theoretical motivation is provided in Appendix \ref{app:hl-loss}. 
Crucially, histogram distributions outputed by $\bar{\mathrm{v}}$ can be transformed back and forth into scalar values using using the HL-Gauss transform operators introduced in Appendix H. 

We now describe the overall architecture of the prediction network. 
The shared core comprises two graph convolutionnal layers $\mathbf{\mathcal{G}}^f_{v \rightarrow c}: \mathcal{H} \rightarrow \mathcal{H}$ and $\mathbf{\mathcal{G}}^f_{c \rightarrow v}: \mathcal{H} \rightarrow \mathcal{H}$ followed by an output module $\mathbf{o_m}: \mathcal{H} \rightarrow \mathbb{R}^{n\times d_{h}}$. This module consists in two fully connected layers followed by non-linear activations. Finally, the three output heads are computed by dedicated linear networks: $\mathbf{o_v}: \mathbb{R}^{n\times d_{h}} \rightarrow \mathbb{R}^{m_b}, \,\mathbf{o_p}:\mathbb{R}^{n\times d_{h}} \rightarrow \mathbb{R}^{n}, \,\mathbf{o_b}:\mathbb{R}^{n\times d_{h}} \rightarrow \mathbb{R}^{2}$ which produce the final predictions from the shared feature representation:

    \begin{align*}
        \mathbf{p} : 
            & \left\{
                \begin{array}{ll}
                    \mathcal{H} \rightarrow \mathbb{R}^{m_b} \\
                    \mathbf{p}(\mathbf{x}) = \mathbf{o_p} \circ \mathbf{o_m} \circ \mathcal{G}^f_{c \rightarrow v} \circ \mathcal{G}^f_{v \rightarrow c}(\mathbf{x})
                \end{array}
            \right. \\
        \bar{\mathbf{v}} : 
            & \left\{
                \begin{array}{ll}
                    \mathcal{H} \rightarrow \mathbb{R} \\
                    \bar{\mathbf{v}}(\mathbf{x}) = \text{pool} \circ \mathbf{o_v} \circ \mathbf{o_m} \circ \mathcal{G}^f_{c \rightarrow v} \circ \mathcal{G}^f_{v \rightarrow c}(\mathbf{x})
                \end{array}
            \right. \\
        \mathbf{b} : 
            & \left\{
                \begin{array}{ll}
                    \mathcal{H} \rightarrow \mathbb{R}^2 \\
                    \mathbf{b}(\mathbf{x}) = \text{pool} \circ \mathbf{o_b} \circ \mathbf{o_m} \circ \mathcal{G}^f_{c \rightarrow v} \circ \mathcal{G}^f_{v \rightarrow c}(\mathbf{x})
                \end{array}
            \right.
    \end{align*}

    where $\text{pool}(\cdot)$ designates the mean pooling operation and $m_b$ the number of bins used to partition the support of the value function, see Appendix H for more details.
    
    \subsubsection*{Dynamics network} 
    In MBRL, the agent has reversible access to the environment's dynamics $(\mathcal{T}, \mathcal{R})$, either through a simulator or a learned model. 
    This constitutes one of the main differences between AlphaZero \cite{silver_general_2018} and MuZero \cite{schrittwieser_mastering_2020}: in AlphaZero, the model consists in a perfect board game simulator, whereas in Muzero the model is a neural network $g$ trained jointly with the prediction network $f$. 
    As described in Section 2, our setting adopts a constant reward function $\mathcal{R} = -1$ and a deterministic transition function $\mathcal{T} = \kappa_\rho$ which boils down to the composition of the branching operation $\kappa$, and the node selection policy $\rho$, such that $\kappa_{\rho} = \kappa \circ \rho$. 
    In this work, we adopt an hybrid approach between AlphaZero and MuZero: we train $g$ to replicate the branching operation $\kappa$ and we simulate $\rho = DFS$, such that our overall B\&B model is part learned and part simulation. 
    
    Our dynamic network $g$ comprises two independent heads $\mathbf{g}^l$ and $\mathbf{g}^r$ dedicated respectively to the left and right child nodes.
    In our convention, the left node is the node first visited by the DFS node selection policy, while the right node is the node visited once its sibling node's subtree has been fathomed. 
    We describe the architecture of $\mathbf{g}^l$, which is identical to the one of $\mathbf{g}^r$. The $\mathbf{g}^l$ module is composed of an action embedding module $\mathbf{e}^l_a$, two graph convolutionnal layers $\mathbf{\mathcal{G}}^l_{v \rightarrow c}$, $\mathbf{\mathcal{G}}^l_{c \rightarrow v}$ and an output module $\mathbf{o}^l_{\mathbf{g}}$. The action embedding module $\mathbf{e}^l_a: \mathcal{H} \times \mathcal{A} \rightarrow \mathcal{H}$ only transforms the representation of the variable node in the bipartite graph corresponding to the variable $a$, by feeding it to a 2-layer fully connected neural network. The output of $\mathbf{e}^l_a$ is fed to the convolutionnal block $\mathbf{\mathcal{G}}^l_{v \rightarrow c}$, $\mathbf{\mathcal{G}}^l_{c \rightarrow v}$, and in turn to the output module $\mathbf{o}^l_{\mathbf{g}}: \mathcal{H}^2 \rightarrow \mathcal{H}$, which is also connected to the output of $\mathbf{e}^l_a$ through a residual link. The dynamic network does not predict rewards, although such functionality could be easily added if one wishes to experiment with alternative reward models. Finally $g$ can be summarized as:
    \begin{align*}
        \mathbf{g}^l : 
            & \left\{
                \begin{array}{ll}
                    \mathcal{H} \times \mathcal{A} \rightarrow \mathcal{H} \\
                    \mathbf{g}^l(\mathbf{x}) = \mathbf{o}^l_{\mathbf{g}} \circ (\textbf{Id} + \, \mathcal{G}^l_{c \rightarrow v} \circ \mathcal{G}^l_{v \rightarrow c}) \circ \mathbf{e}^l_a (\mathbf{x}) 
                \end{array}
            \right. \\
        \mathbf{g}^r : 
            & \left\{
                \begin{array}{ll}
                    \mathcal{H} \times \mathcal{A} \rightarrow \mathcal{H} \\
                    \mathbf{g}^r(\mathbf{x}) = \mathbf{o}^r_{\mathbf{g}} \circ (\textbf{Id} + \, \mathcal{G}^r_{c \rightarrow v} \circ \mathcal{G}^r_{v \rightarrow c}) \circ \mathbf{e}^r_a (\mathbf{x}) 
                \end{array}
            \right. \\
    \end{align*}

    \paragraph{Simulating subtree trajectories} Let $\hat{o} \in \mathcal{H}$ be the internal representation of the B\&B current node $o = \rho(s_t)$ and let $a \in \mathcal{A}$ be an action to perform. The dynamics network $g$ generates $\hat{o}_l, \hat{o}_r$ the internal representations of the two child nodes created when branching on variable $a$ at $s_t$: $\hat{o}_l, \hat{o}_r = g(\hat{o}, a)$. The policy, subtree value and branchability associated with $\hat{o}_l, \hat{o}_r$ are computed using the prediction network: $\mathrm{p}_{\hat{o}_l}, \bar{\mathrm{v}}_{\hat{o}_l}, {\mathrm{b}}_{\hat{o}_l}= f(\hat{o}_l) \, ; \, \mathrm{p}_{\hat{o}_r}, \bar{\mathrm{v}}_{\hat{o}_r}, {\mathrm{b}}_{\hat{o}_r}= f(\hat{o}_r)$.
    Then, the two child nodes are added to the imagined tree $\hat{\mathrm{T}} =\{\hat{\mathcal{O}}, \hat{\mathcal{C}}\}$ based on their predicted branchability. Unsurprisingly, if $\hat{o}_i \in \{ \hat{o}_l, \hat{o}_r\}$ is predicted to be branchable, it is added to the set of imagines open nodes $\hat{\mathcal{O}}$, otherwise, it is added to the set of imagined closed nodes $\hat{\mathcal{C}}$. 
    Crucially, the predictions $\mathrm{p}_{\hat{o}_i}, \bar{\mathrm{v}}_{\hat{o}_i}, {\mathrm{b}}_{\hat{o}_i}$ are stored in $\hat{\mathrm{T}}$ along with $\hat{o}_i$ to enable future computations of the imagined tree's associated policy and state value. 
    For example, let $\hat{\mathrm{T}}^k = (\hat{\mathcal{O}}^k, \hat{\mathcal{C}}^k )$ be the imagined subtree obtained after unrolling the PlanB\&B model for $k$ step from $s_t$. Our node selection policy simulator selects the next node to expand $\tilde{o} = \rho(\hat{\mathcal{O}}^k)$ as the node in $\hat{\mathcal{O}}^k$ with the greatest depth. In case of equality, the next current node is the left most node in $\hat{\mathcal{O}}$ with highest depth. This way, our model is compatible with any DFS priority criterion discriminating between nodes of equal depth. As highlighted in Section 3, the estimate policy $\hat{\mathrm{p}}^k_t$ and state value $\hat{\mathrm{v}}^k_t$ associated with $\hat{\mathrm{T}}^k$ can be retrieved from $\hat{\mathcal{O}}^k$ through the formulas: $
    \hat{\mathrm{p}}^k_t = \mathrm{p}_{\tilde{o}}  \; ; \; \hat{\mathrm{v}}^k_t = \sum_{\hat{o} \in \hat{\mathcal{O}}^k} \bar{\mathrm{v}}_{\hat{o}}$. 
    To simplify notations, in the following sections we abusively write $\hat{\mathrm{p}}^k_t, \hat{\mathrm{v}}^k_t = f(\hat{\mathrm{T}}^k)$ and $\hat{\mathrm{T}}^{k+1} = g(\hat{\mathrm{T}}^k, a)$ to summarize the global interaction between $f$ and $g$ to simulate subtree trajectories.

\section*{C \; Search}
\label{app:mcts}
We describe the planning algorithm used in PlanB\&B. Our approach builds on Gumbel Search, a variant of Monte Carlo Tree Search (MCTS) that provides stronger policy improvement guarantees, particularly in settings with limited simulation budgets.
In PlanB\&B, every node of the search tree is associated with an imagined B\&B subtree $\hat{\mathrm{T}}$ rooted in the current B\&B node $o = \rho(s_t)$. For each action $a \in \mathcal{A}$ available at node $\hat{\mathrm{T}}$, there is a corresponding edge storing the statistics $N(\hat{\mathrm{T}},a), Q(\hat{\mathrm{T}},a), P(\hat{\mathrm{T}},a)$  respectively representing the visit count, normalized $Q$-value and policy prior associated with the pair $(\hat{\mathrm{T}},a)$. As in MCTS, the search algorithm is divided in three stages: path selection, search tree expansion and value backpropagation, repeated for a number $N$ of simulations. In Gumbel search, only the path selection procedure differs from traditional MCTS implementations.

\paragraph{Selection} Each simulation consists in a path $(a^1, \ldots, a^l)$ starting from the root of the search tree, represented by the initial subtree $\hat{\mathrm{T}}^0 = (\mathcal{O}^0, \mathcal{C}^0)$ where $\mathcal{O}^0 = \{\hat{o}\}$ and $\mathcal{C}^0 = \emptyset$, and diving towards a leaf node $\hat{\mathrm{T}}^l$ that is yet to be expanded. To generate such paths, Gumbel Search uses two different types of selection criterion for choosing $a^k$ for $k=1\ldots l$, depending on whether $\hat{\mathrm{T}}^{k-1}$ correspond to the root of the search tree or to a deeper node. 
In fact, as emphasized by \citet{danihelka2022policy}, since the goal is to derive a policy improvement procedure via planning, the search algorithm should aim not to minimize cumulative regret over the $N$ simulations, as is the case in traditional MCTS implementations, but rather the simple regret associated with the action selected at the root of the search tree at iteration $N + 1$. Accordingly, at the root node, our search algorithm employs the Sequential Halving procedure \citep{karnin2013almost} in combination with the Gumbel-Top-$k$ trick \citep{kool2019stochastic}, as described by \citet{danihelka2022policy}, to efficiently sample a promising subset of $M < |\mathcal{A}|$ actions for exploration. This procedure is particularly well suited to the MILP setting, as it preserves the policy improvement property even when the available simulation budget is small relative to the size of the action set. In contrast, beyond root node, $a^k$ is simply selected as :

\begin{equation*}
     a^k = \underset{a \in \mathcal{A}}{\arg \max} \left( \pi'(\hat{\mathrm{T}}^{k-1}, a) - \frac{N(\hat{\mathrm{T}}^{k-1}, a)}{1+\sum_{b\in \mathcal{A}} N(\hat{\mathrm{T}}^{k-1}, b)}\right)
\end{equation*}

where $\pi'(\hat{\mathrm{T}}^{k-1}, a)$ is a function of $Q(\hat{\mathrm{T}}^{k-1}, a)$ and $P(\hat{\mathrm{T}}^{k-1}, a)$ introduced in the next paragraphs.
% To generate such paths, action $a^0, a^1, \ldots, a^l$ are selected following a exploration-exploitation criterion. At iteration $l=0$, 

\paragraph{Expansion}
 At the final step of the simulation, upon reaching the unvisited edge $(\hat{\mathrm{T}}^{l-1}, a^l)$, the transition $\hat{\mathrm{T}}^l = g(\hat{\mathrm{T}}^{l-1}, a^l)$ is computed by applying the dynamics network. The policy and value estimates for $\hat{\mathrm{T}}^l$ are then obtained via the prediction network such that  $\hat{\mathrm{p}}^l, \hat{\mathrm{v}}^l = f(\hat{\mathrm{T}}^l)$. The search tree is subsequently expanded by initializing each outgoing edge  $(\hat{\mathrm{T}}^l, a)$ with the statistics: $\{N(\hat{\mathrm{T}}^l,a) = 0, Q(\hat{\mathrm{T}}^l,a) = 0, P(\hat{\mathrm{T}}^l,a) = \hat{\mathrm{p}}^l\}$. Importantly, each simulation requires a single call to the dynamics function, and two calls to the prediction network.

\paragraph{Backpropagation}
Once the search tree is expanded, the statistics $N$ and $Q$ are updated along the simulation path, starting from $\hat{\mathrm{T}}^l$. For $k= l \ldots 1$, noting $G^ k= -(l - k) + \hat{\mathrm{v}}^l$, each edge $(\hat{\mathrm{T}}^{k-1}, a^k)$ is updated following :

\begin{align*}
    Q(\hat{\mathrm{T}}^{k-1}, a^k)  &:= \frac{N(\hat{\mathrm{T}}^{k-1}, a^k) \cdot Q(\hat{\mathrm{T}}^{k-1}, a^k) + G^k}{N(\hat{\mathrm{T}}^{k-1}, a^k) +1}\\
    N(\hat{\mathrm{T}}^{k-1}, a^k)  &:= N(\hat{\mathrm{T}}^{k-1}, a^k) + 1\\
\end{align*}

\paragraph{Improved policy target}

After completing $N$ simulation steps, an improved policy target $\pi_t$ is computed from the statistics accumulated at the root of the search tree, such that $\pi_t(a) = \pi'(\hat{\mathrm{T}}^0, a)$ for ${a \in \mathcal{A}}$, with $\pi'$ defined as : 
\begin{align*}
    \pi'(\hat{\mathrm{T}}^k, a) &= \text{softmax}(P(\hat{\mathrm{T}}^k, a) + \sigma(\tilde{Q}(\hat{\mathrm{T}}^k, a))).
\end{align*}

with $\sigma(\tilde{Q}(\hat{\mathrm{T}}^k, a))) = (c_{visit} + \underset{b\in \mathcal{A}}{\max} N(\hat{\mathrm{T}}^k, b)) \cdot c_{scale} \cdot \tilde{Q}(\hat{\mathrm{T}}^k, a))$. 
In turn, $\tilde{Q}(\hat{\mathrm{T}}^k, a))$ is defined as :

\[ \tilde{Q}(\hat{\mathrm{T}}^k, a)) = 
    \left\{
        \begin{array}{ll}
            Q(\hat{\mathrm{T}}^k, a) \; \; \; \; \; \; \; \; \;\text{if} \; N(\hat{\mathrm{T}}^k, a) > 0 \\
             \sum P(\hat{\mathrm{T}}^k, a) \cdot Q(\hat{\mathrm{T}}^k, a) \; \; \; \; \; \; \text{else}. \\

        \end{array}
    \right.
\]

Similar to the policy targets generated by standard MCTS procedures, the improved Gumbel Search policies achieve a trade-off between exploration and exploitation to guide action selection. Importantly, as in MuZero, the $Q$-value statistics used to balance exploration and exploitation are in fact normalized values $Q^\dagger \in [0, 1]$, computed using a simple normalization scheme applied across the search tree:

\begin{align*}
    &Q^\dagger(\hat{\mathrm{T}}^{k-1}, a^k) = \\ &\frac{Q(\hat{\mathrm{T}}^{k-1}, a^k) - \min_{(\hat{\mathrm{T}}, a) \in Tree } Q(\hat{\mathrm{T}},a)}{\max_{(\hat{\mathrm{T}}, a) \in Tree } Q(\hat{\mathrm{T}},a) - \min_{(\hat{\mathrm{T}}, a) \in Tree} Q(\hat{\mathrm{T}},a)}.
\end{align*}

\section*{D \; Strong Branching}
\label{app:milp_planing}
In this section, we present strong branching (SB), the branching expert used in \citet{gasse_exact_2019}, and show how it can be reformulated as a MCTS procedure when considering a dual bound maximization objective. 

\paragraph{Strong branching} Among the various branching heuristics proposed in the literature, strong branching stands out as one of the most powerful and effective, albeit computationally expensive, techniques. The core idea of SB is to simulate the effect of branching on several candidate variables before actually committing to a decision. This is achieved by tentatively imposing the two branching disjunctions associated with each candidate variable (e.g., $x_i \leq \lfloor \hat{x}_i \rfloor$ and $x_i \geq \lceil \hat{x}_i \rceil$ for fractional variable $x_i$) and solving the corresponding child nodes' LP relaxations to obtain dual bounds.
The SB score of a variable is typically based on the estimated dual bound improvement, that is, how much the LP relaxation’s objective increases under each branching direction. Intuitively, a variable is promising for branching if both child nodes yield significantly higher associated dual bounds $c^Tx^*_{LP}$ than the current node, indicating that branching on this variable may help prune the search tree more effectively. 
%Multiple scoring schemes have been proposed to aggregate the improvements from both branches, such as the minimum, average, or weighted sum of improvements.
Mathematically, for a candidate variable $x_i$, let $\Delta_l^i$ and $\Delta_r^i$ denote the increases in the LP relaxation objective value under the left and right branch respectively. A common scoring function for SB writes 
$\text{score}(x_i) = \min(\Delta^i_l, \Delta^i_r)$. Such scoring schemes aims to identify variables that exhibit balanced improvement in both branches, thus maximizing the likelihood of early node fathoming.

Despite its effectiveness in reducing the size of generated B\&B trees, the primary drawback of strong branching is its computational overhead. At each node, it requires solving potentially hundreds of LPs, one for each branching direction of every fractional variable, which quickly becomes prohibitive even for medium-scale MILPs. As a result, modern solvers typically reserve strong branching for early tree nodes or use it in hybrid strategies, where a small number of variables are pre-selected based on cheaper scoring functions, and S is applied selectively.
In recent years, S has also served as a supervised learning target for data-driven branching strategies. Due to its high computational cost but high accuracy, its decisions provide a valuable signal for training imitation learning models that seek to replicate expert decisions at a fraction of the computational expense \citep{gasse_exact_2019}.
In summary, strong branching represents a gold standard for variable selection in B\&B, as it produces best known expert branching decisions. Its study continues to inform both theoretical analyses of B\&B efficiency and the design of advanced branching heuristics, including those based on learning.

\paragraph{Strong Branching as Lookahead Planning} 
Strong branching can be naturally interpreted as a one-step Monte Carlo Tree Search (MCTS) procedure aimed at maximizing the dual bound. In this perspective, the current B\&B node corresponds to the root of the search tree, and each candidate variable $x_i$ defines a branching action $a_i$ leading to a leaf node associated with an expanded B\&B tree. For each action $a_i$, strong branching simulates both branches by solving the LP relaxations of the left and right child nodes, thus obtaining dual bounds $z^i_l$ and $z^i_r$. The aggregated score $\text{score}(x_i) = \min(z^i_l, z^i_r)$ serves as an estimate of the value of the branching decision $a_i$, and is propagated back to the root of the search tree to guide variable selection.

Crucially, strong branching differs from traditional MCTS in that it performs a full-width, depth-one expansion of all available actions without search tree rollouts or simulations. It estimates action quality based purely on immediate lookahead, and selects the action with the best estimated dual bound improvement. This can be interpreted as an extremely expensive form of planning under a one-step horizon, where the dynamics model is replaced by LP solvers, and the value function corresponds to the dual bound improvement. In this light, strong branching serves as a handcrafted, domain-specific instantiation of value-based planning, motivating the exploration of more general model-based planning approaches for improving variable selection in B\&B. 

%This reformulation also clarifies the connection between classical B\&B heuristics and recent learning-based planning methods. Just as AlphaZero and MuZero refine action selection by simulating multiple futures and updating policies through bootstrapped rollouts, strong branching approximates a single-step lookahead with perfect evaluation, thus offering a computationally expensive but informative target for imitation and model-based reinforcement learning.

\section*{E \; Learning} 
\label{app:loss}

We describe the overall training mechanism of PlanB\&B. Crucially, PlanB\&B is trained over $K$-step subtree trajectories $(s_t, a_t, ..., s_{t+K})$. As illustrated in Figure~\ref{fig:MuZero}, the PlanB\&B model is unrolled along the same action sequence as the historical sampled trajectory, generating a sequence of imagined subtrees $(\hat{\mathrm{T}}^0, \ldots, \hat{\mathrm{T}}^K)$. Along these subtree trajectories, the policy network $\mathbf{p}$ is trained to replicate $\pi_{t+k}$ the policy outputed by the search algorithm, while the subtree value network $\bar{\mathbf{v}}$ is trained to match $z_{t+k}$, the $n$-step return target. As in MuZero, our value network is trained via classification, using the HL-Gauss loss introduced in Appendix H. Moreover, in PlanB\&B the branchability network $\mathbf{b}$ is trained to discriminate between branchable and unbranchable nodes.
Losses used to train each network are described by the following equations:
\begin{align*}
    \hat{\mathrm{p}}^k_{t} &= \{ \mathbf{p}_{\hat{o}} \, | \, \hat{o} = \rho(\hat{\mathcal{O}}^k) \} \\
    \hat{\mathrm{v}}^k_{t} &= \sum_{\hat{o} \in \hat{\mathcal{O}}^{k}} \bar{\mathbf{v}}_{\hat{o}} \\
    \pi_{t+k} &= Gumbel(s_{t+k}, h, f, g \; ; \theta) \\
    z_{t+k} &= -n + \hat{\mathrm{v}}^{k+n}_t \\
    \mathcal{L}_\mathrm{p}(\pi, \hat{\mathrm{p}}) &= \mathbf{\pi}^\top \log \hat{\mathrm{p}} \\
    \mathcal{L}_\mathrm{v}(z, \hat{\mathrm{v}}) &= z^{\top} \log \hat{\mathrm{v}}\\
    \mathcal{L}_\mathrm{b}(\mathrm{b}, \hat{\mathrm{b}}) &= \mathrm{b}^{\top} \log \hat{\mathrm{b}}\\
\end{align*}

\paragraph{Tree consistency loss} Following \citet{ye_mastering_2021}, we introduce a new self-supervised loss $\mathcal{L}_\mathrm{t}$ to enforce tree consistency between consecutive internal B\&B node representations.
% Let $\hat{\mathrm{T}}_{k} = (\hat{\mathcal{O}}_k, \hat{\mathcal{C}}_k)$ be the imagined tree associated with $s_{t+k}$, and $\hat{o} \in \mathcal{H}$ be the internal representation associated with $o = \rho(s_{t})$ the current B\&B node. 
Let $\hat{o} \in \mathcal{H}$ be the internal representation associated with $o = \rho(s_{t})$ the current B\&B node. 
By definition, $\hat{o} = h(o, \bar{x}_{t})$.
When feeding $\hat{o}$ to the dynamics network $g$ along with $a_{t}$, we obtain $\hat{o}_l, \hat{o}_r$, the internal representation of the child nodes associated with branching on $a_{t}$ at $s_{t}$. Let us write $\tau_l, \tau_r$ the time steps at which the real nodes $o_l$, $o_r$ associated with $\hat{o}_l$, $\hat{o}_r$ are visited.
Provided that these time steps are finite, we can compute the latent representation targets $h(o_l, \bar{x}_{\tau_l}), h(o_r, \bar{x}_{\tau_r}) \in \mathcal{H}$ for $\hat{o}_l$ and $\hat{o}_r$. Otherwise, $o_i$ is unbranchable, in that case we refrain from enforcing tree consistency.

\begin{figure}[t]
    \centering
    \includegraphics[width=0.30\textwidth]{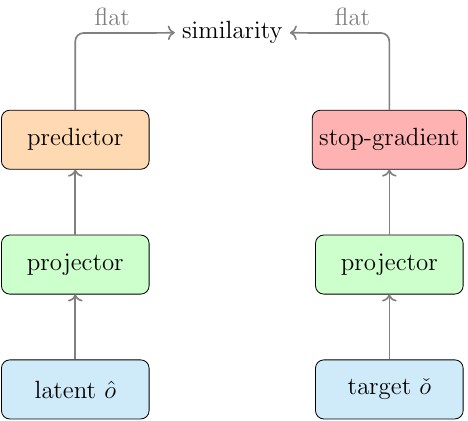}
    \caption{PlanB\&B tree-consistency loss modeled after the SimSiam architecture from \cite{chen_exploring_2020}.}
    \label{fig:self-supervised}
\end{figure}

Our self-supervised loss $\mathcal{L}_t$ takes inspiration from the SimSiam architecture from \citet{chen_exploring_2020}. Let $\hat{o} \in \mathcal{H}$ be the internal representation of node $o \in \mathcal{O}_{t+k}$ obtained after unrolling the PlanB\&B model for $k$ steps, and let $\check{o} \in \mathcal{H}$ be its associated target representation as defined in the previous paragraph. As illustrated in Figure \ref{fig:self-supervised}, given $\mathbf{r}_{proj}$ and $\mathbf{r}_{pred}$ two fully connected block modules with associated hidden dimension $d_{proj}$, the self-supervised consistency loss writes:

\begin{equation*}
    \mathcal{L}_{\mathrm{t}}({\check{o}, \hat{o}}) = \text{sim}(\text{flat} \circ \text{sg} \circ \mathbf{r}_{proj}(\check{o}), \text{flat} \circ \mathbf{r}_{pred} \circ \mathbf{r}_{proj}(\hat{o})) \\
\end{equation*}

with $\text{flat}(\cdot)$ the flattening operation, $\text{sg}(\cdot)$ the stop-gradient operation and $\text{sim}(\cdot)$ the cosine similarity. Thus, the overall training objective associated with a trajectory $(s_t, a_t, ..., s_{t+K})$ is given by:

\begin{align}
    \label{eq:full_loss}
    \mathcal{L}_t(\theta) &= \frac{1}{K+1}\sum_{k=0}^{K} \left[ \lambda_{\mathrm{p}} \mathcal{L}_\mathrm{p}(\pi_{t+k}, \hat{\mathrm{p}}^k_{t}) + \lambda_{\mathrm{v}}\mathcal{L}_\mathrm{v}(z_{t+k}, \hat{\mathrm{v}}^k_{t}) \right] \nonumber \\
    & + \frac{1}{|\hat{\mathrm{T}}^K|} \sum_{\hat{o} \in \hat{\mathrm{T}}^K} \left[ \lambda_{\mathrm{b}} \mathcal{L}_{\mathrm{b}}(\mathrm{b}_{o}, \hat{\mathrm{b}}_{\hat{o}}) + \lambda_{\mathrm{t}}\mathcal{L}_{\mathrm{t}}({\check{o}, \hat{o}}) \right] 
\end{align}

with $\lambda_{\mathrm{p}}, \lambda_{\mathrm{v}}, \lambda_{\mathrm{b}}$ and $\lambda_{\mathrm{t}}$ hyper-parameter loss weights.

\section*{F \; Training pipeline}
\label{app:rl}
\begin{figure}[t]
    \centering
    \includegraphics[width=0.49\textwidth]{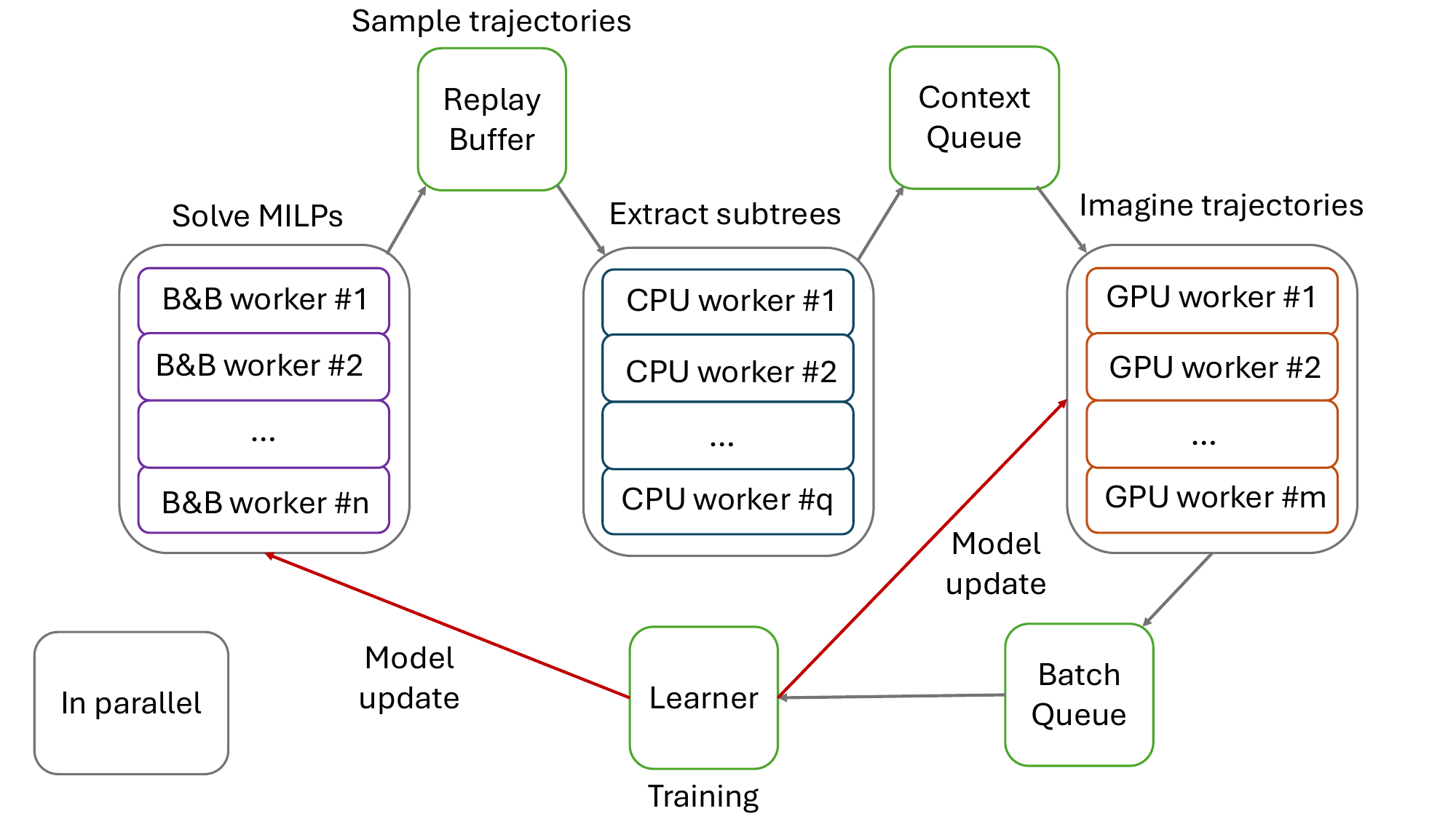}
    \caption{PlanB\&B training pipeline, build upon the framework from \citet{ye_mastering_2021}.}
    \label{fig:pipeline}
\end{figure}

Our PlanB\&B implementation builds on the official implementation from \citet{ye_mastering_2021} and \citet{wang2024efficientzero}. Table~\ref{tab:hyperparams} reports the hyperparameter values that were modified to suit our training setting.
Our training pipeline follows the parallelized architecture introduced in EfficientZero, implementing a double-buffering mechanism, as illustrated in Figure \ref{fig:pipeline}. The training process operates synchronously across multiple components:
\begin{itemize}
    \item MILP actors run B\&B episodes using the current model, updated every 100 training steps, to produce policy targets $\pi_t$ via Gumbel Search and push generated trajectories into a shared replay buffer. 
    \item CPU rollout workers sample these trajectories from replay buffer, and preprocess them by extracting B\&B subtree trajectories from full B\&B episodes, an operation confined to CPU resources.
    \item GPU batch workers unroll the PlanB\&B model for $K$ step using the target model, executing the most compute-intensive steps on the GPU.
    \item Finally, the learner worker receives both historic and imagined subtree trajectories and performs gradient updates according to Eq. \eqref{eq:full_loss}.
\end{itemize}
All components run in parallel. The replay buffer is shared between MILP actors and CPU workers, while a context queue is used for communication between CPU and GPU workers. A separate batch queue connects GPU workers and the learner. This design ensures efficient utilization of both CPU and GPU resources throughout training. Importantly, all workers except for CPU workers leverage GPU acceleration to evaluate the PlanB\&B model. All experiments were conducted on a server equipped with an Intel(R) Xeon(R) Platinum 8480CL 128-core processor, 1024 GB of RAM and 4×NVIDIA A100 GPUs (40GB VRAM each). Resources were efficiently allocated across all workers using the Ray library, ensuring optimal workload distribution. 

\begin{table}[t]
    \small
    \centering
    \begin{tabular}{lc} 
        Parameter &  Setting \\
        \toprule
         Training steps & $10^5$ \\
         Batch size & $128$ \\
         Optimizer & Adam \\
         Learning rate $l_r$ & $10^{-3} \rightarrow 10^{-5}$ \\
         Model unroll step ($K$) & $3$ \\
         TD steps ($n$) & $3$ \\
         Discount factor $\gamma$ & $1.0$ \\
         Policy loss coefficient $\lambda_{\mathrm{p}}$ & $1$ \\
         Value loss coefficient $\lambda_{\mathrm{v}}$ & $1$ \\
         Branchability loss coefficient $\lambda_{\mathrm{b}}$ & $1$ \\
         Tree consistency loss coefficient $\lambda_{\mathrm{t}}$ & $1$ \\
         HL-Gauss min log value $z_{min}$ & -1 \\
         HL-Gauss max log $z_{max}$ & $16$ \\
         HL-Gauss number of bins $m_b$ &$18$ \\
         HL-Gauss $\sigma_G$ & $0.75$  \\
         Number of simulations $N$ & $50$ \\
         Gumbel search root node action number $M$ & $10$ \\
         Gumbel search shift factor $c_{visit}$ & $50$ \\
         Gumbel search scaling factor $c_{scale}$ & $0.1$ \\
         Variable node feature dimension $d_v$ & $43$ \\
         Constraint node feature dimension $d_c$ & $5$ \\
         Edge feature dimension $d_c$ & $1$ \\
         Latent space embedding dimension $d_{h}$ & $64$ \\
         Projection space dimension $d_{proj}$ & $16$ \\
         Replay buffer capacity & $10^5$ \\
         MILP solving time limit (s)  & $3600$ \\
        \bottomrule
    \end{tabular}
    \caption{Training parameters for PlanB\&B. Crucially, missing hyperparameter settings were kept fixed as in \citet{wang2024efficientzero} implementation.}
\label{tab:hyperparams}
\end{table}

\section*{G \; Baselines}
\label{app:baseline}

\paragraph{Imitation learning} We trained and tested IL agents using the official implementation of \citet{gasse_exact_2019} shared at \url{https://github.com/ds4dm/learn2branch-ecole/tree/main}.

\paragraph{DQN-TreeMDP} Since there is no publicly available implementation of \citet{etheve_reinforcement_2020}, we re-implemented DQN-TreeMDP and trained it on the four Ecole benchmarks, using exactly the same network architectures and training parameters as in DQN-Retro.

\paragraph{PG-tMDP} We used the official implementation of \citet{scavuzzo_learning_2022} to evaluate PG-TreeMDP. For each benchmark, we used the tMDP+DFS network weights shared at \url{https://github.com/lascavana/rl2branch}.

\paragraph{DQN-Retro} As \citet{parsonson_reinforcement_2022} only trained on set covering instances, we took inspiration from the official implementation shared at \url{https://github.com/cwfparsonson/retro_branching} to train and evaluate the DQN-Retro agent on the four Ecole benchmarks. Crucially, we trained and tested DQN-Retro following SCIP default node selection strategy. 

\paragraph{}Importantly, on the multiple knapsack benchmark, given the high computational cost associated with opting for a depth-first search node selection policy, all baselines excepted IL-DFS are evaluated following SCIP default node selection policy.

\section*{H \; HL-Gauss Loss}
\label{app:hl-loss}
As they investigated the uneven success met by complex neural network architectures such as Transformers in supervised versus reinforcement learning, \citet{farebrother_stop_2024} found that training agents using a cross-entropy classification objective significantly improved the performance and scalability of value-based RL methods. However, replacing mean squared error regression with cross-entropy classification requires methods to transform scalars into distributions and distributions into scalars. \citet{farebrother_stop_2024} found the Histogram Gaussian loss (HL-Gauss) \citep{imani_improving_2018}, which exploits the ordinal structure structure of the regression task by distributing probability mass on multiple neighboring histogram bins, to be a reliable solution across multiple RL benchmarks. Concretely, in HL-Gauss, the support of the value function $\mathcal{Z} \subset \mathbb{R}$ is divided in $m_b$ bins of equal width forming a partition of $\mathcal{Z}$. Bins are centered at $z_i \in \mathcal{Z}$ for ${1 \leq i \leq m_b}$, we use $\eta = (z_{max}-z_{min})/m_b$ to denote their width. Given a scalar $z \in \mathcal{Z}$, we define the random variable $Y_z \sim \mathcal{N}(\mu=z, \sigma_G^2)$ and note respectively $\phi_{Y_z}$ and $\Phi_{Y_z}$ its associate probability density and cumulative distribution function. $z$ can then be encoded into a histogram distribution on $\mathcal{Z}$ using the function $p_{hist}: \mathbb{R} \rightarrow [0, 1]^{m_b}$. Explicitly, $p_{hist}$ computes the aggregated mass of $\phi_{Y_z}$ on each bin as $p_{hist}(z) = (p_i(z))_{1\leq i \leq m_b}$ with:

\[p_i(z) = \int_{z_i-\frac{\eta}{2}}^{z_i+\frac{\eta}{2}} \phi_{Y_z}(y) dy = \Phi_{Y_z}(z_i+\frac{\eta}{2}) - \Phi_{Y_z}(z_i-\frac{\eta}{2}). \]

\begin{figure*}[t]
\centering
\begin{subfigure}{0.48\textwidth}
    \centering
    \includegraphics[width=\linewidth]{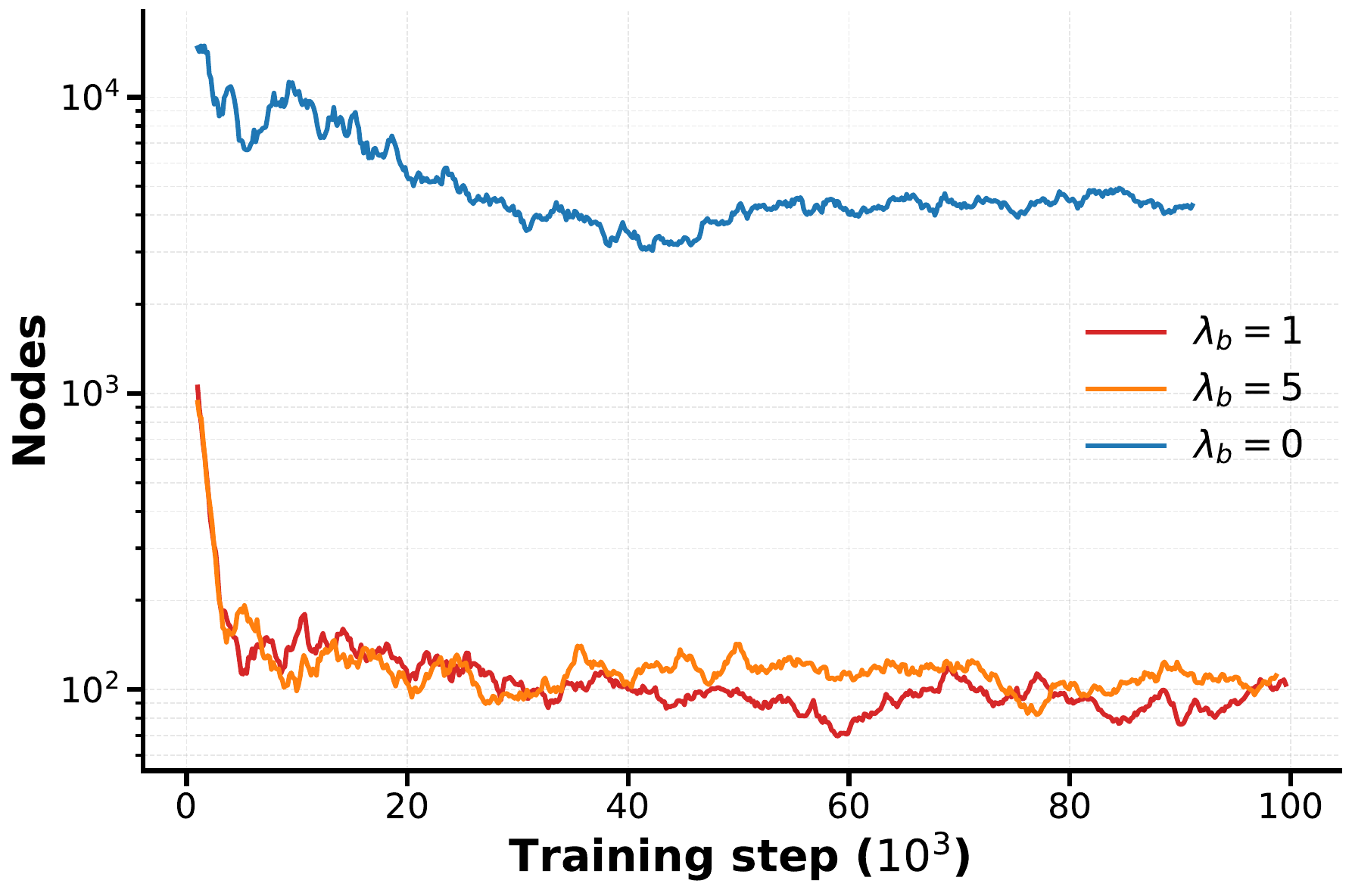}
    \caption{Tree size validation}
    \label{fig:breward_performance}
\end{subfigure}
%\vspace{1em}  % Adds vertical spacing between the subfigures
\begin{subfigure}{0.48\textwidth}
    \centering
    \includegraphics[width=\linewidth]{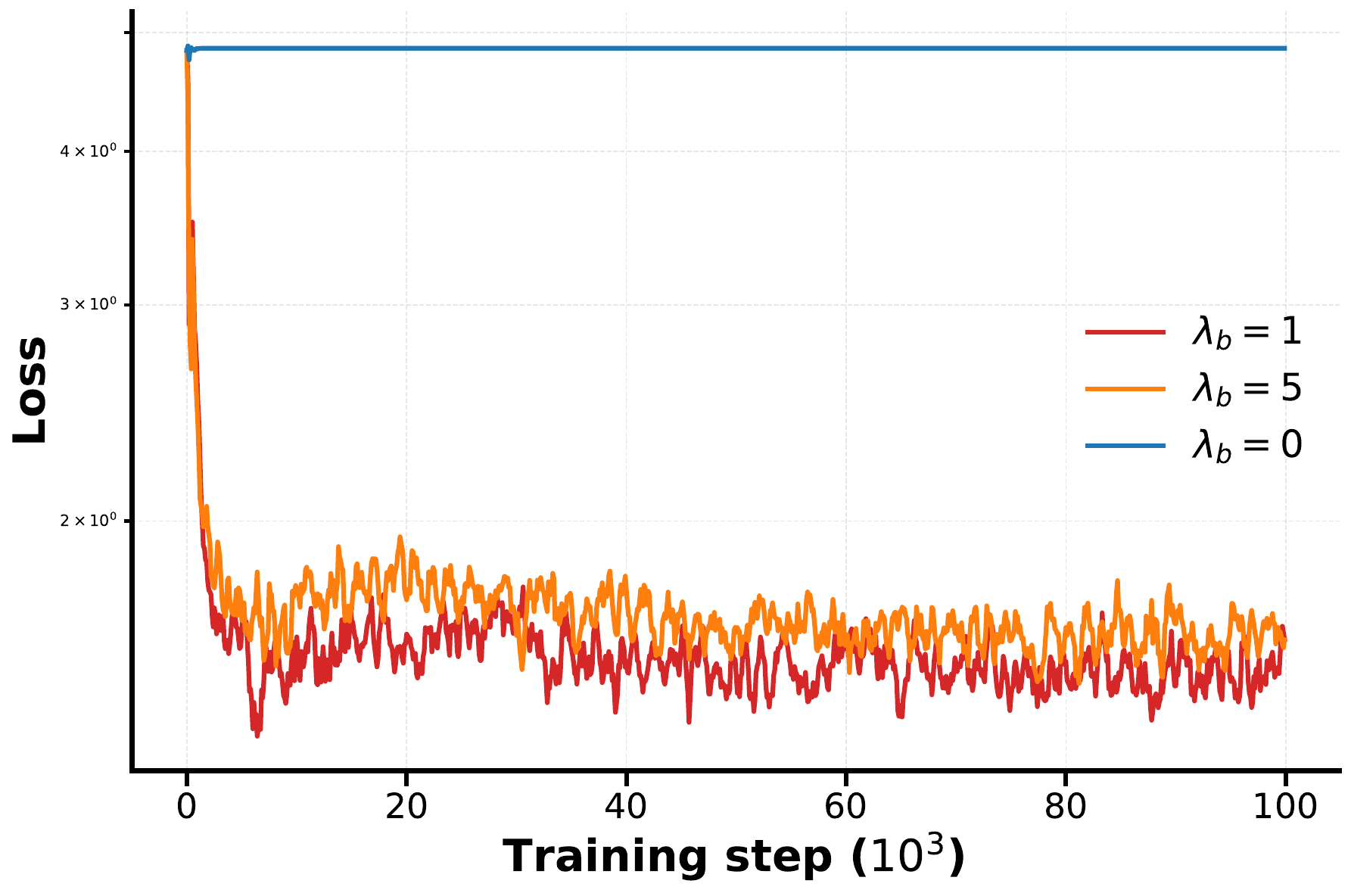}
    \caption{Branchability loss}
    \label{fig:breward_loss}
\end{subfigure}
\begin{subfigure}{0.48\textwidth}
    \centering
    \includegraphics[width=\linewidth]{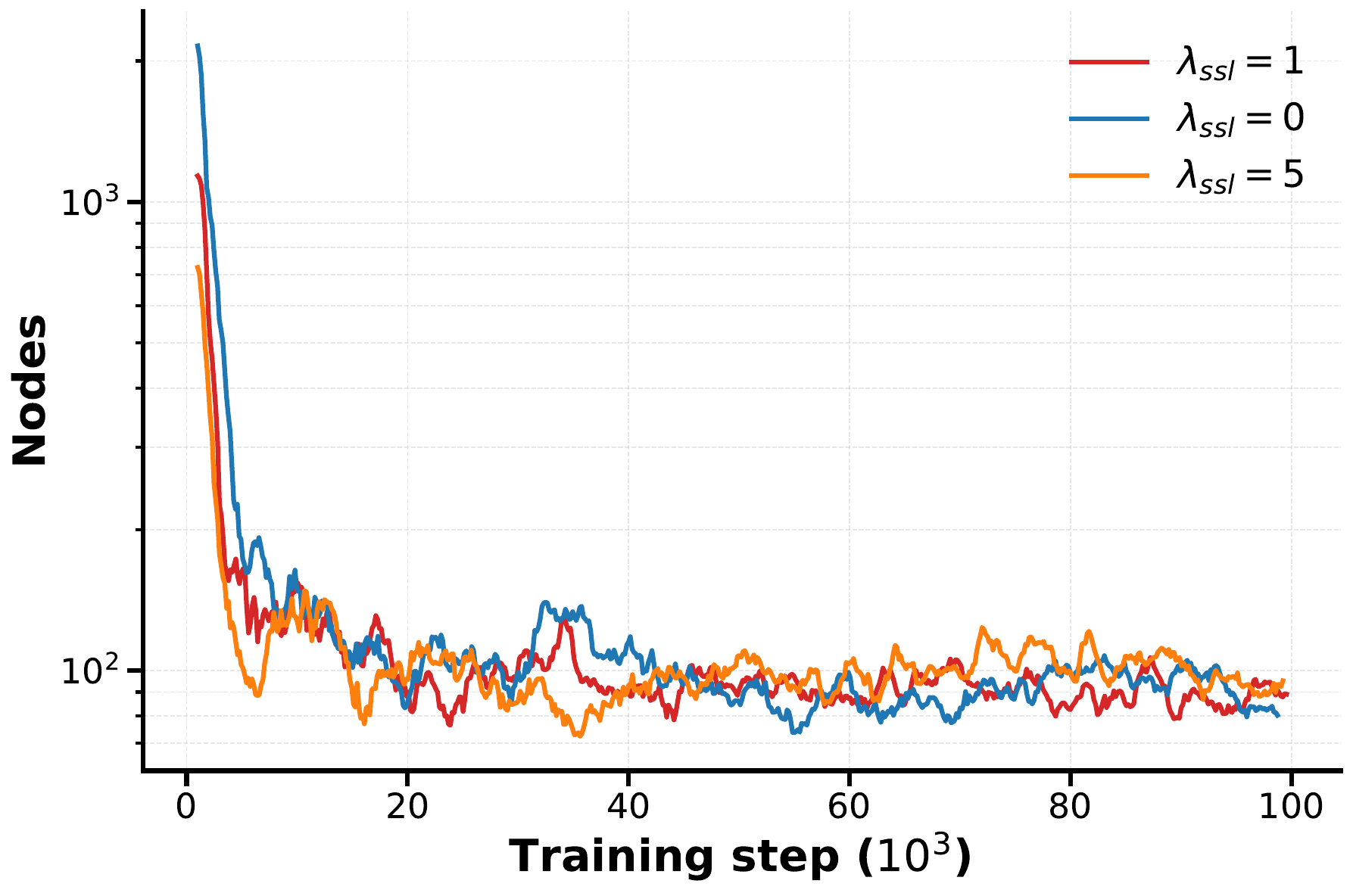}
    \caption{Tree size validation}
    \label{fig:ssl_performance}
\end{subfigure}
%\vspace{1em}  % Adds vertical spacing between the subfigures
\begin{subfigure}{0.48\textwidth}
    \centering
    \includegraphics[width=\linewidth]{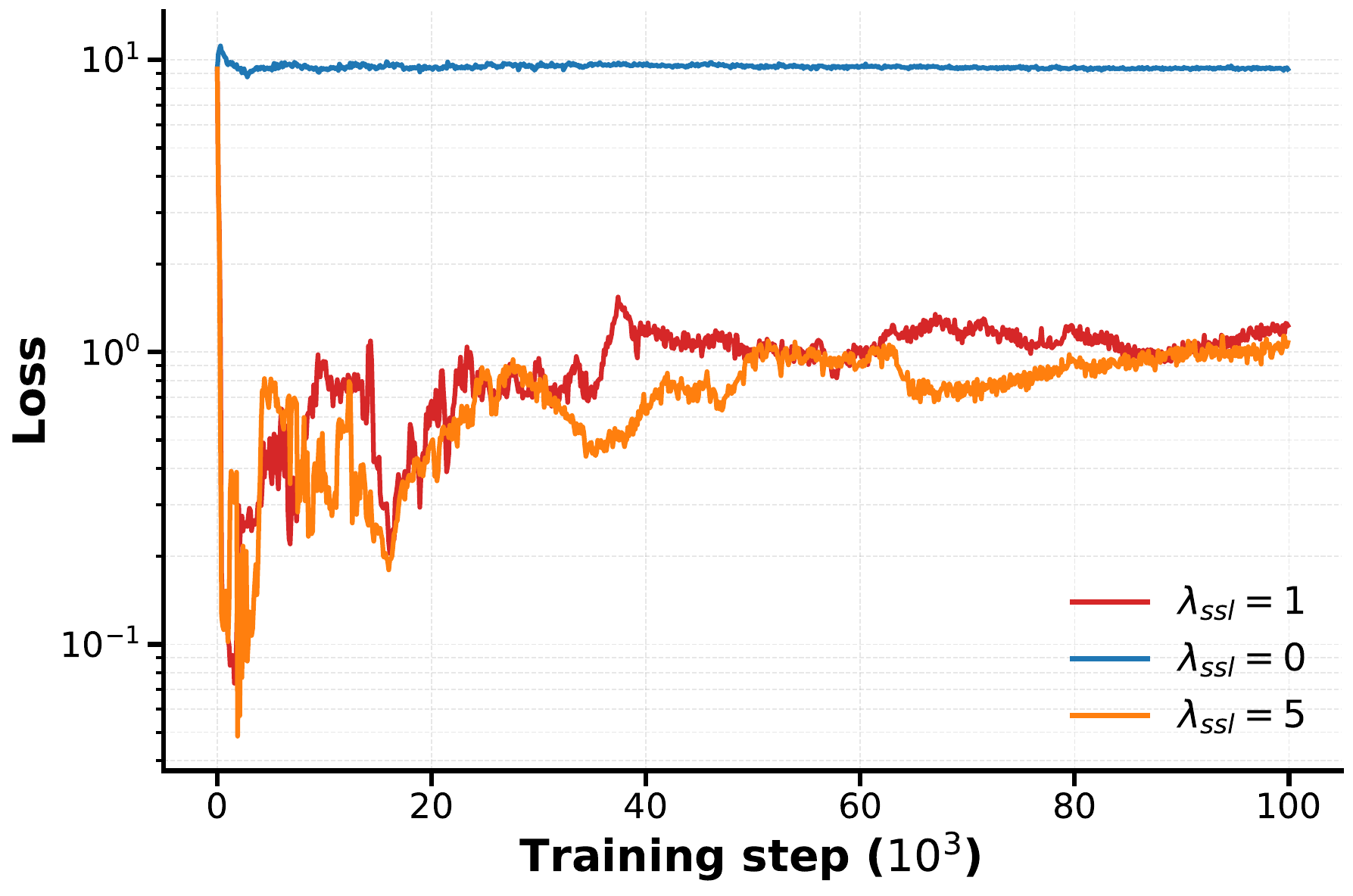}
    \caption{Temporal consistency loss}
    \label{fig:ssl_loss}
\end{subfigure}

\caption{PlanB\&B ablation study for branchability and temporal consistency losses on combinatorial auction (CA) instances.}
\label{fig:targeted_ablations}
\end{figure*}

Conversely, histogram distributions $(p_i)_{1 \leq i \leq m_b}$ such as the ones outputted by agents' value networks can be converted to scalar simply by computing the expectation: $z = \sum_{i=1}^{m_b} p_i \cdot z_i$. \\
\newline PlanB\&B is a challenging setting to adapt HL-Gauss, as the support for value functions spans over several order of magnitude. In practice, we observe that for train instances of the Ecole benchmark, $\mathcal{Z} = [-10^6, -1]$. Since value functions predict the number of node of binary trees built with B\&B, it seems natural to choose bins centered at $z_i = -2^{i}$ to partition $\mathcal{Z}$. In order to preserve bins of equal size, we consider distributions on the support $\psi(\mathcal{Z})$ with $\psi(z) = \log_2(-z)$ for $z \in \mathcal{Z}$, such that $\psi(\mathcal{Z})$ is efficiently partitioned by bins centered at $z_i=i$ for ${0 \leq i \leq m_b}$.  Thus, histograms distributions are given by $p_{hist}(z) = (p_i \circ \psi(z))_{1 \leq i \leq m_b}$ for $z \in \mathcal{Z}$, and can be converted back to $\mathcal{Z}$ through $z = \sum_{i=0}^{m_b} p_i \cdot \psi^{-1}(z_i)$ with $\psi^{-1}(z) = -2^{z}$.

\begin{figure*}[t]
    \centering
    \begin{subfigure}{0.48\textwidth}
        \centering
        \includegraphics[width=\linewidth]{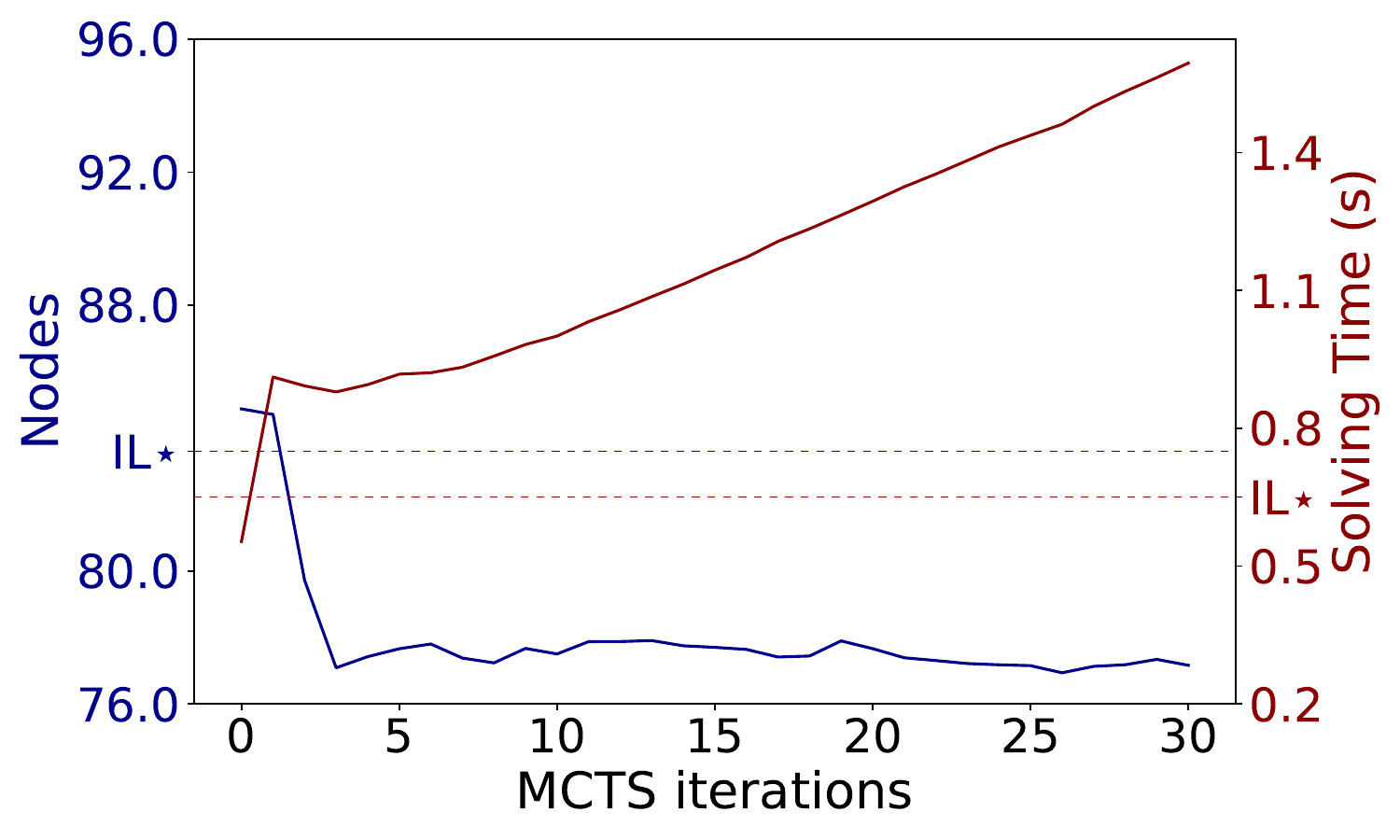}
        \caption{Test instances}
        \label{fig:chart1}
    \end{subfigure}
    %\vspace{1em}  % Adds vertical spacing between the subfigures
    \begin{subfigure}{0.48\textwidth}
        \centering
        \includegraphics[width=\linewidth]{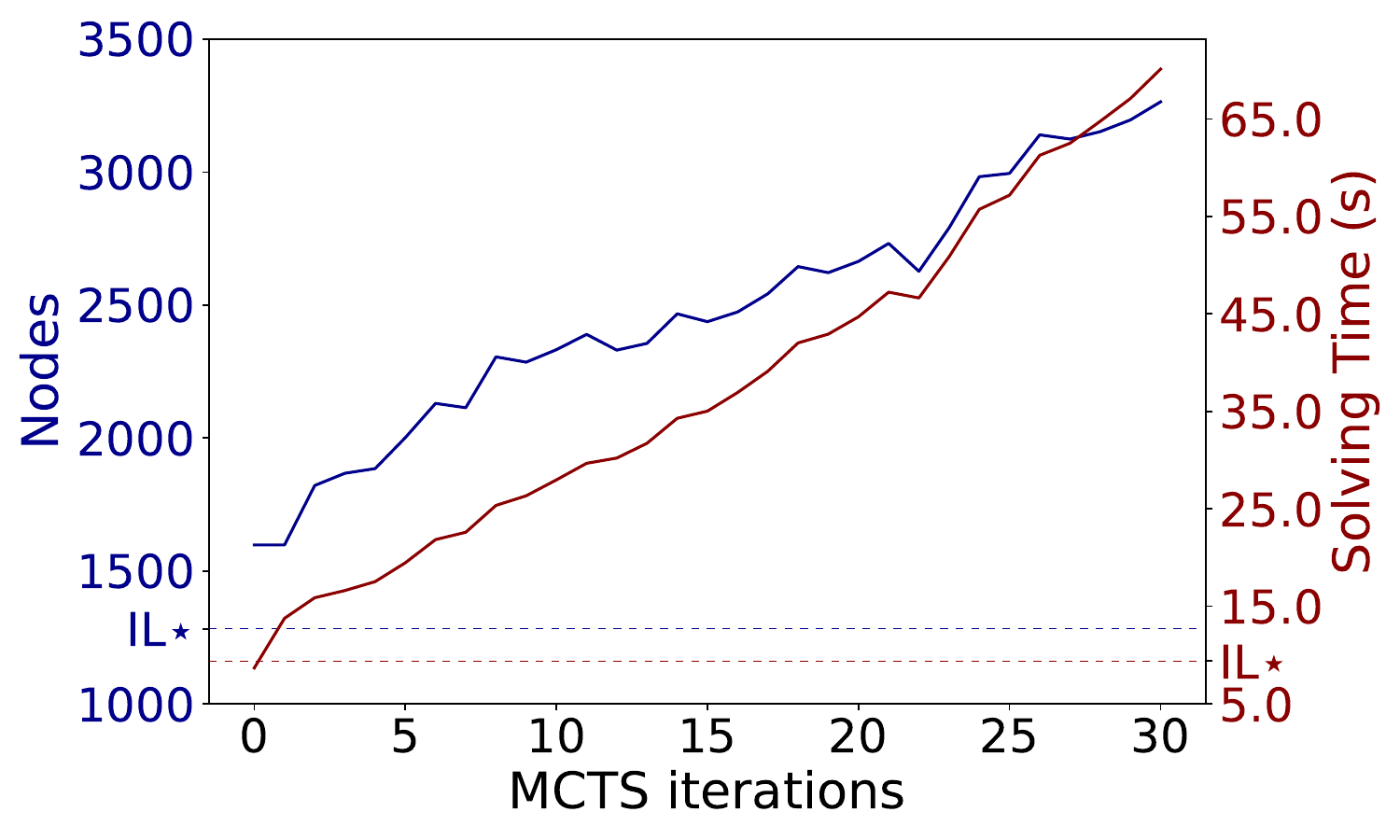}
        \caption{Transfer instances}
        \label{fig:chart2}
    \end{subfigure}

    \caption{Policy improvement associated with increased simulation budget over the CA benchmark.}
    \label{fig:mcts_combined_ca}
\end{figure*}

\section*{I \; Further computational results}
In this section, we provide additional background to support and complement the claims presented in Section 5.

\paragraph{Targeted ablations (Q1)} 
We investigate how the branchability and temporal consistency losses introduced in PlanB\&B affect the agent’s convergence behavior and final performance. 
Ablation results, including training curves and validation metrics on combinatorial auction instances, are reported in Figure~\ref{fig:targeted_ablations}. First, as shown in Figures \ref{fig:breward_performance}-\ref{fig:breward_loss}, learning to distinguish between branchable and unbranchable nodes is indispensable for the agent to be able to improve its current policy through planning. 
When setting $\lambda_{\mathrm{b}}=0$, the agent fails to derive any benefit from search, because it no longer reliably detects episode termination, and therefore cannot assign meaningful values to simulated rollouts.
As a result, the branchability loss emerges as a key ingredient for PlanB\&B to translate search into steady performance gains.

Second, our results suggest that the temporal consistency loss provides, at best, a modest acceleration of training, and has little measurable impact on final performance under our data regime. As shown in Figure~\ref{fig:ssl_performance}, any advantage from setting $\lambda_{\mathrm{ssl}} > 0$ is short-lived, largely vanishing after the first $5$--$10$k training steps. This is consistent with the training loss reported in Figure~\ref{fig:ssl_loss}: after a rapid initial decrease over the first few thousand steps, the temporal consistency loss gradually rises again before settling at a relatively high plateau. More generally, this behavior matches the intended role of temporal-consistency objectives in MBRL: they primarily act as an auxiliary shaping signal that improves representation learning and stabilizes optimization early on, while long-run performance is ultimately governed by the accuracy of the policy, value, and branchability predictions. In sum, while our experiments suggest that the self-supervised consistency loss is not a decisive ingredient for PlanB\&B on the Ecole benchmark, it may prove more valuable in more challenging regimes (e.g., larger instances), where stabilizing training and speeding up early learning become critical.

\paragraph{Additional performance metrics (Q1)} Table \ref{tab:win_results} provides additional performance metrics to compare the different baselines across test and transfer instances. For each benchmark, we report the number of wins and the average rank of each baseline across 100 evaluation instances. The number of wins is defined as the number of instances where a baseline solves a MILP problem faster than any other baseline. When multiple baselines fail to solve an instance to optimality within the time limit, their performance is ranked based on the final dual gap. Finally, Table \ref{tab:complete_results} recapitulates the computational results presented in Table 1, and provides for each baseline the per-benchmark standard deviation over five seeds, as well as the fraction of test instances solved to optimality within the time limit. 

\paragraph{Value network generalization (Q2)} To contextualize the results presented in Section 5, we provide additional discussion on the impact of leveraging the PlanB\&B model to improve branching decisions. In our experiments, we observe that outside of the MIS benchmark, the value head $\bar{\mathbf{v}}$ of the prediction network $f$ struggles to generalize to higher-dimensional MILPs. We attribute this to out-of-distribution effects: the value network is trained on instances with structurally smaller subtrees than those encountered in the transfer benchmarks. 
Similar limitations are observed in prior RL baselines: PG-tMDP demonstrates stronger generalization than both DQN agents, likely due to its reliance on a policy network rather than a $Q$-network to make branching decisions.
Fortunately, this limitation does not prevent the policy network $\mathbf{p}$ from producing efficient branching strategies on transfer instances, as shown in Table 1. However, it does hinder the ability of PlanB\&B to refine branching decisions at inference time using its learned model, as the inaccuracies of the value predictions undermine planning. This is illustrated in Figure \ref{fig:mcts_combined_ca} on combinatorial auction instances: improved node performance can be achieved by leveraging model-based planning on test instances, but not on transfer instances. This limitation could potentially be addressed in future work by gradually increasing the dimensionality of the instances solved during training by the MILP actors, thereby allowing the value network to progressively adapt to larger B\&B trees such as the one encountered in transfer benchmarks.

\paragraph{Strong branching alignment metrics (Q3)} Table 2 reports three metrics designed to quantify the alignment between learned policies and the strong branching (SB) expert over MIS instances.
First, we report the average cross-entropy between the predicted policy and the SB policy, normalized by the cross-entropy between the SB policy and a uniform random policy (SB C-Entropy).
Second, we report the average strong branching score of the action selected by the learning baseline, normalized by the SB score of the action selected by the SB expert (SB Score).
Third, we report the average frequency with which the learning baseline selects the same action as the SB expert (SB Freq).
All metrics are averaged over 100 test instances and 20 higher-dimensional transfer instances.

\paragraph{Influence of DFS on B\&B solving time (Q4)} The overall impact of adopting a depth-first search (DFS) node selection policy is more nuanced than initially suggested in Section 5. In fact, as previously argued, DFS generally yields larger branch-and-bound (B\&B) trees compared to the advanced node selection heuristics used in modern MILP solvers, which often results in longer solving times. However, DFS also introduces a notable computational advantage: when nodes are explored along diving trajectories, solvers can leverage warm-start techniques to accelerate the resolution of successive linear programs. This speed-up can be quite substantial, as illustrated in Table~\ref{tab:DFS2}.
As a result, the net performance impact of DFS arises from the interplay between these two opposing effects: larger trees that increase computational workload, and faster LP solves that help mitigate this cost. This trade-off helps explain part of the performance gaps observed in Table 1 between DFS and non-DFS baselines. 
In particular, it clarifies how PlanB\&B manages to solve test instances faster than the ${\text{IL}^\star}$ baseline despite producing larger trees in average. Nonetheless, DFS remains suboptimal overall when compared to the more advanced node selection strategies implemented in modern solvers.

\begin{table}[t]
    \centering
    \small
    \begin{tabular}{ccccc}
    & \multicolumn{4}{c}{Test instances} \\
    & Set Cov. & Comb. Auct. & Max. Ind. Set & Mult. Knap. \\
    \toprule
        SCIP & $11.0$ & $21.4$ & $12.2$ & $22.0 $ \\
        DFS & $7.2 $ & $18.3$ & $8.7$ & $19.9$ \\
        \midrule
        Gain & $-30\%$ & $-10\%$ & $-25\%$ & $-14\%$ \\
    \bottomrule
    \vspace{0.1mm}
    \end{tabular}
    \begin{tabular}{ccccc}
    & \multicolumn{4}{c}{Transfer instances} \\
   & Set Cov. & Comb. Auc. & Max. Ind. Set & Mult. Knap. \\
    \toprule
        SCIP & $16.3$ & $12.4$ & $36.0 $ & $4.3 $ \\
        DFS & $11.8$ & $9.3$ & $28.1$ & $3.6$ \\
        \midrule
        Gain & $-25\%$ & $-25\%$ & $-22\%$ & $-16\%$ \\
    \bottomrule
     
    \end{tabular}
    \caption{Average transition time (ms) associated with the execution of $\kappa_{\rho}$ across the Ecole benchmark. We evalute both SCIP's default node selection policy and depth-first search. }
    \label{tab:DFS2}
\end{table}

    \begin{table*}[t]
        \centering
        \small
        \begin{tabular}{ccccccc} 
            & \multicolumn{6}{c}{Set Covering} \\
            & \multicolumn{3}{c}{Train / Test} & \multicolumn{3}{c}{Transfer}\\
            Method & Solved & Wins & Rank & Solved &  Wins & Rank   \\  
            \toprule
            SCIP  & $100/100$ & $3/100$ & $5.73$ & $100/100$ & $2/100$ & $4.31$ \\
            \midrule
             ${\text{IL}^\star}$ & $100/100$ & $0/100$ & $3.53$ & $100/100$ & $29/100$ & $1.85$ \\
             IL-DFS & $100/100$ & $33/100$ & $\textcolor{purple}{\mathbf{2.10}}$ & $100/100$ & $\textcolor{purple}{\mathbf{58/100}}$ & $\textcolor{purple}{\mathbf{1.70}}$ \\
            \midrule
            PG-tMDP  & $100/100$ & $0/100$ & $6.6$ & $78/100$ & $0/100$ & $6.95$ \\
            DQN-tMDP  & $100/100$ & $12/100$ & $2.74$ & $96/100$ & $1/100$ & $4.92$ \\ 
            DQN-Retro & $100/100$ & $0/100$ & $4.67$ & $99/100$ & $0/100$ & $5.12$ \\
            PlanB\&B & $100/100$ & $\textcolor{purple}{\mathbf{52/100}}$ & $\textcolor{blue}{\mathbf{2.76}}$ & $100/100$ & $\textcolor{blue}{\mathbf{10/100}}$ & $\textcolor{blue}{\mathbf{3.17}}$ \\
            \bottomrule
        \hspace{1.5cm}
        \end{tabular}
            \begin{tabular}{ccccccc} 
            & \multicolumn{6}{c}{Combinatorial Auction} \\
            & \multicolumn{3}{c}{Train / Test} & \multicolumn{3}{c}{Transfer}\\
            Method & Solved & Wins & Rank & Solved &  Wins & Rank  \\ 
            \toprule
            SCIP  & $100/100$ & $4/100$ & $6.16$ & $100/100$ & $5/100$ & $3.86$ \\
            \midrule
             ${\text{IL}^\star}$ & $100/100$ & $10/100$ & $3.3$ & $100/100$ & $28/100$ & $\textcolor{purple}{\mathbf{2.12}}$\\
             IL-DFS & $100/100$ & $26/100$ & $2.7$ & $100/100$ & $26/100$ & $2.47$  \\
            \midrule
            PG-tMDP  & $100/100$ & $2/100$ & $5.08$ & $100/100$ & $0/100$ & $5.83$\\
            DQN-tMDP  & $100/100$ & $2/100$ & $5.08$ & $100/100$ & $0/100$ & $9.97$ \\ 
            DQN-Retro & $100/100$ & $7/100$ & $3.47$ & $100/100$ & $1/100$ & $4.63$ \\
            PlanB\&B & $100/100$ & $\textcolor{purple}{\mathbf{42/100}}$ & $\textcolor{purple}{\mathbf{2.6}}$ &  $100/100$ & $\textcolor{purple}{\mathbf{40/100}}$ & $\textcolor{purple}{\mathbf{2.12}}$ \\
            \bottomrule
        \hspace{1.5cm}
        \end{tabular}
        \begin{tabular}{ccccccc} 
            & \multicolumn{6}{c}{Maximum Independent Set} \\
            & \multicolumn{3}{c}{Train / Test} & \multicolumn{3}{c}{Transfer}\\
            Method & Solved & Wins & Rank & Solved &  Wins & Rank \\  
            \toprule
            SCIP  & $100/100$ & $2/100$ & $5.72$ & $100/100$ & $5/100$ & $4.40$ \\
            \midrule
             ${\text{IL}^\star}$ & $100/100$ & $38/100$ & $\textcolor{purple}{\mathbf{2.21}}$ & $100/100$ & $\textcolor{purple}{\mathbf{33/100}}$ & $\textcolor{purple}{\mathbf{2.11}}$ \\
             IL-DFS & $100/100$ & $14/100$ & $2.9$ & $100/100$ & $0/100$ & $3.23$\\
            \midrule
            PG-tMDP  & $100/100$ & $0/100$ & $4.62$ & $100/100$ & $23/100$ & $3.1$ \\
            DQN-tMDP  & $100/100$ & $0/100$ & $4.68$ & $85/100$ & $0/100$ & $6.06$ \\ 
            DQN-Retro & $100/100$ & $2/100$ & $5.51$ & $22/100$ & $0/100$ & $6.84$\\
            PlanB\&B & $100/100$ & $\textcolor{purple}{\mathbf{44/100}}$ & $\textcolor{blue}{\mathbf{2.45}}$ & $100/100$ & $\textcolor{blue}{\mathbf{32/100}}$ & $\textcolor{blue}{\mathbf{2.26}}$\\
            \bottomrule
        \hspace{1.5cm}
        \end{tabular}
        \begin{tabular}{ccccccc} 
            & \multicolumn{6}{c}{Multiple Knapsack} \\
            & \multicolumn{3}{c}{Train / Test} & \multicolumn{3}{c}{Transfer}\\
            Method & Solved & Wins & Rank & Solved &  Wins & Rank   \\  
            \toprule
            SCIP  & $\textcolor{purple}{\mathbf{100/100}}$ & $\textcolor{purple}{\mathbf{73/100}}$ & $\textcolor{purple}{\mathbf{1.60}}$ & $\textcolor{purple}{\mathbf{100/100}}$ & $\textcolor{purple}{\mathbf{53/100}}$ & $\textcolor{purple}{\mathbf{2.11}}$ \\
            \midrule
             ${\text{IL}^\star}$ & $100/100$ & $0/100$ & $4.60$ & $100/100$ & $7/100$ & $3.39$\\
             IL-DFS & $100/100$ & $2/100$ & $5.83$ & $100/100$ & $0/100$ & $6.11$ \\
            \midrule
            PG-tMDP  & $100/100$ & $0/100$ & $6.03$ & $98/100$ & $4/100$ & $5.19$ \\
            DQN-tMDP  & $100/100$ & $1/100$ & $3.71$ & $99/100$ & $11/100$ & $5.71$ \\ 
            DQN-Retro & $100/100$ & $3/100$ & $3.61$ & $100/100$ & $9/100$ & $4.01$ \\
            PlanB\&B & $\textcolor{blue}{\mathbf{100/100}}$ & $\textcolor{blue}{\mathbf{21/100}}$ & $\textcolor{blue}{\mathbf{2.63}}$ & $\textcolor{blue}{\mathbf{100/100}}$ & $             \textcolor{blue}{\mathbf{16/100}}$ & $\textcolor{blue}{\mathbf{3.48}}$ \\
            \bottomrule
        \hspace{1cm}
        \end{tabular}
        \caption{Additional performance metrics for each baseline on train / test and transfer instance benchmarks, see Appendix A for instance details. For each benchmark, we report the number of wins, and the average rank of each baseline across the 100 evaluation instances. We also report for each baseline the fraction of test instances solved to optimality within time limit. The number of wins is defined as the number of instances where a baseline solves a MILP problem faster than all other baselines. When multiple baselines fail to solve an instance to optimality within time limit, their performance is ranked based on dual gap. }
        \label{tab:win_results}
    \end{table*}

\label{app:results}
    \begin{table*}[t]
        \centering
        \small
        \begin{tabular}{ccccccc} 
            & \multicolumn{3}{c}{Train / Test} &  \multicolumn{3}{c}{Transfer}\\
            Method & Nodes & Time & Solved & Nodes & Time & Solved \\ 
            \toprule
            Random & $3289 \pm 4.2\%$ & $5.94 \pm 4.3\%$ & $100/100$ &  $271632 \pm 12.7\%$ & $842 \pm 9.8\%$ & $60/100$ \\
            SB  & $35.8 \pm 0.0\%$ & $12.93 \pm 0.0\%$ & $100/100$ &  $672.1 \pm 0.0\%$ & $398 \pm 0.2\%$ & $82/100$ \\
            SCIP  & $62.0 \pm 0.0\%$ & $2.27 \pm 0.0\%$ & $100/100$ & $3309 \pm 0.0\%$ & $48.4 \pm 0.1\%$ & $100/100$ \\
            \midrule
             ${\text{IL}^\star}$ & \textcolor{purple}{$\mathbf{133.8 \pm 1.0\%}$} & $0.90 \pm 4.8\%$ & $100/100$ & \textcolor{purple}{$\mathbf{2610 \pm 0.7\%}$} & $23.1 \pm 1.5\%$ & \textcolor{purple}{$\mathbf{100/100}$} \\
             IL-DFS & $136.4 \pm 1.8\%$ & \textcolor{purple}{$\mathbf{0.74 \pm 5.3\%}$} & $100/100$ & $3103 \pm 2.0\%$ & \textcolor{purple}{$\mathbf{22.5 \pm 3.1\%}$} & \textcolor{purple}{$\mathbf{100/100}$} \\
            \midrule
            PG-tMDP  & $649.4 \pm 0.7\%$ & $2.32 \pm 2.4\%$ & $100/100$ & $44649 \pm 3.7\%$ & $221 \pm 4.1\%$ & $78   /100$ \\
            DQN-tMDP  & \textcolor{blue}{$\mathbf{175.8\pm 1.1\%}$} & \textcolor{blue}{$\mathbf{0.83 \pm 4.5\%}$} & $100/100$ & $8632 \pm 4.9\%$ & $71.3 \pm 5.8\%$ & $96/100$ \\ 
            DQN-Retro & $183.0 \pm 1.2\%$ & $1.14 \pm 4.1\%$ & $100/100$ & $6100 \pm 4.2\%$ & $59.4 \pm 4.2\%$ & $98/100$ \\
            PlanB\&B & $186.2 \pm 0.4\%$ & $0.87 \pm 6.0\%$ & $100/100$ & \textcolor{blue}{$\mathbf{5869\pm 2.5\%}$} & \textcolor{blue}{$\mathbf{46.2 \pm 3.1\%}$} & \textcolor{blue}{$\mathbf{100/100}$} \\
            \bottomrule
            & \multicolumn{6}{c}{Set covering}
        \hspace{1cm}
        \end{tabular}
        \begin{tabular}{ccccccc} 
            & \multicolumn{3}{c}{Train / Test} & \multicolumn{3}{c}{Transfer}\\
            Method & Nodes & Time & Solved & Nodes & Time & Solved \\  
            \toprule
            Random & $1111 \pm 4.3\%$ & $2.16 \pm 6.6\%$ & $100/100$ & $3172355 \pm 7.5\%$ & $749 \pm 9.1\%$ & $64/100$ \\
            SB  & $28.2 \pm 0.0\%$ & $6.21 \pm 0.1\%$ & $100/100$ & $389.6 \pm 0.0\%$ & $255 \pm 0.2\%$ & $88/100$ \\
            SCIP  & $20.2 \pm 0.0\%$ & $1.77 \pm 0.1\%$ & $100/100$ & $1376 \pm 0.0\%$ & $14.
            77\pm 0.1\%$ & $100/100$ \\
            \midrule
             ${\text{IL}^\star}$ & \textcolor{purple}{$\mathbf{83.6 \pm 0.8\%}$} & $0.65 \pm 8.5\%$ & $100/100$ & \textcolor{purple}{$\mathbf{1282 \pm 1.6\%}$} & $9.4 \pm 2.0\%$ & \textcolor{purple}{$\mathbf{100/100}$} \\
             IL-DFS & $95.5 \pm 0.9\%$ & $0.56 \pm 7.2\%$ & $100/100$ & $1828 \pm 2.0\%$ & $10.2 \pm 1.6\%$ & $100/100$ \\
            \midrule
            PG-tMDP  & $168.0 \pm 2.8\%$ & $0.94 \pm 6.0\%$ & $100/100$ & $6001 \pm 2.7\%$ & $30.7 \pm 2.4\%$ & $100/100$ \\
            DQN-tMDP  & $203.3 \pm 4.2\%$ & $1.11 \pm 4.0\%$ & $100/100$ & $20553 \pm 3.8\%$ & $116 \pm 3.9\%$ & $100/100$ \\ 
            DQN-Retro & $103.2 \pm 1.2\%$ & $0.78 \pm 7.5\%$ & $100/100$ & $2908 \pm 1.7\%$ & $18.4 \pm 2.7\%$ & $100/100$ \\
            PlanB\&B & \textcolor{blue}{$\mathbf{84.7 \pm 1.4\%}$} & \textcolor{purple}{$\mathbf{0.54 \pm 7.9\%}$} & $100/100$ & \textcolor{blue}{$\mathbf{1665 \pm 2.3\%}$} & \textcolor{purple}{$\mathbf{9.1 \pm 2.4\%}$} & \textcolor{purple}{$\mathbf{100/100}$} \\
            \bottomrule
            & \multicolumn{6}{c}{Combinatorial auction}
        \hspace{5mm}
        \end{tabular}
                \begin{tabular}{ccccccc} 
            & \multicolumn{3}{c}{Train / Test} & \multicolumn{3}{c}{Transfer}\\
            Method & Nodes & Time & Solved & Nodes & Time & Solved \\ 
            \toprule
            Random & $386.8 \pm 5.4\%$ & $2.01 \pm 4.8\%$ & $100/100$ & $215879 \pm 6.7\%$ & $2102 \pm 6.2\%$ & $25/100$ \\
            SB  & $24.9 \pm 0.0\%$ & $45.87 \pm 0.4\%$ & $100/100$ & $169.9 \pm 0.2\%$ & $2172 \pm 0.9\%$ & $15/100$ \\
            SCIP  & $19.5 \pm 0.0\%$ & $2.44 \pm 0.4\%$ & $100/100$ & $3368 \pm 0.0\%$ & $90.0 \pm 0.2\%$ & $100/100$ \\
            \midrule
             ${\text{IL}^\star}$ & \textcolor{purple}{$\mathbf{40.1 \pm 3.45\%}$} & $0.36 \pm 3.4\%$ & $100/100$ & \textcolor{purple}{$\mathbf{1882 \pm 4.0\%}$} & \textcolor{purple}{$\mathbf{38.6 \pm 3.3\%}$} & \textcolor{purple}{$\mathbf{100/100}$} \\
             IL-DFS & $68.5 \pm 6.5\%$ & $0.44 \pm 4.1\%$ & $100/100$ & $3501 \pm 2.7\%$ & $51.9 \pm 2.1\%$ & $100/100$ \\
            \midrule
            PG-tMDP  & $153.6 \pm 5.0\%$ & $0.92 \pm 2.6\%$ & $100/100$ & $3133 \pm 4.6\%$ & $43.6 \pm 2.9\%$ & $100/100$ \\
            DQN-tMDP  & $168.0 \pm 5.6\%$ & $1.00 \pm 3.4\%$ & $100/100$ & $45634 \pm 7.4\%$ & $477 \pm 5.1\%$ & $85/100$ \\ 
            DQN-Retro & $223.0 \pm 4.1\%$ & $1.81 \pm 3.6\%$ & $100/100$ & $119478 \pm 6.1\%$ & $1863 \pm 4.8\%$ & $22/100$ \\
            PlanB\&B & \textcolor{blue}{$\mathbf{44.8 \pm 7.6\%}$} & \textcolor{purple}{$\mathbf{0.32 \pm 6.4\%}$} & $100/100$ & \textcolor{blue}{$\mathbf{2853 \pm 4.9\%}$} & \textcolor{blue}{$\mathbf{41.1 \pm 5.4\%}$} & \textcolor{blue}{$\mathbf{100/100}$} \\
            \bottomrule
            & \multicolumn{6}{c}{Maximum independent set}
        \hspace{1cm}
        \end{tabular}
        \begin{tabular}{ccccccc} 
            & \multicolumn{3}{c}{Train / Test} & \multicolumn{3}{c}{Transfer}\\
            Method & Nodes & Time & Solved & Nodes & Time & Solved \\ 
            \toprule
            Random & $733.5 \pm 13.0\%$ & $0.55 \pm 6.9\%$ & $100/100$ & $93452\pm 14.3\%$ & $70.6 \pm 9.2\%$ & $99/100$ \\
            SB  & $161.7 \pm 0.0\%$ & $0.69\pm 0.1\%$ & $100/100$ & \textcolor{purple}{$\mathbf{1709 \pm 0.5\%}$} & \textcolor{purple}{$\mathbf{12.5 \pm 0.9\%}$} & \textcolor{purple}{$\mathbf{100/100}$} \\
            SCIP  & $289.5 \pm 0.0\%$ & $0.53 \pm 0.2\%$ & $100/100$ & $30260 \pm 0.0\%$ & $22.1 \pm 0.2\%$ & $100/100$ \\
            \midrule
             ${\text{IL}^\star}$ & $272.0 \pm 12.9\%$ & $0.69 \pm 8.1\%$ & $100/100$ & $11730 \pm 7.1\%$ & $43.5 \pm 6.4\%$ & $100/100$ \\
             IL-DFS & $411.5 \pm 13.0\%$ & $1.07 \pm 8.8\%$ & $100/100$ & $43705 \pm 9.2\%$ & $130.8 \pm 8.3\%$ & $98/100$ \\
            \midrule
            PG-tMDP  & $436.9 \pm 21.2\%$ & $1.57 \pm 16.9\%$ & $100/100$ & $35614 \pm 14.3\%$ & $123 \pm 15.4\%$ & $98/100$ \\
            DQN-tMDP  & $266.4 \pm 7.2\%$ & $0.73 \pm 4.6\%$ & $100/100$ & $22631 \pm 8.6\%$ & $65.1 \pm 5.5\%$ & $99/100$ \\ 
            DQN-Retro & $250.3 \pm 9.5\%$ & $0.67 \pm 5.0\%$ & $100/100$ & $27077 \pm 8.8\%$ & $79.5 \pm 6.2\%$ & $100/100$ \\
            PlanB\&B & \textcolor{purple}{$\mathbf{220.0 \pm 5.9\%}$} & \textcolor{purple}{$\mathbf{0.55 \pm 4.9\%}$} & $100/100$ & \textcolor{blue}{$\mathbf{13574 \pm 6.6\%}$} & \textcolor{blue}{$\mathbf{51.2 \pm 4.8\%}$} & \textcolor{blue}{$\mathbf{100/100}$} \\
            \bottomrule
            & \multicolumn{6}{c}{Multiple knapsack}
        \hspace{5mm}
        \end{tabular}
        \caption{Computational performance comparison on four MILP benchmarks. Following prior works, we report geometrical mean over 100 instances, averaged over 5 seeds, as well as per-benchmark standard deviations.}
        \label{tab:complete_results}
    \end{table*}

\end{document}